\documentclass[preprint,12pt,authoryear]{elsarticle}

\usepackage{amssymb}
\usepackage{amsmath}
\numberwithin{equation}{section}
\usepackage[table]{xcolor}
\usepackage{multirow}
\usepackage{subcaption}
\usepackage{booktabs}
\usepackage{algorithm}
\usepackage{algpseudocode}
\usepackage{pdflscape}
\usepackage{url}
\usepackage{soul}
\usepackage{mathrsfs}
\usepackage{verbatim}
\usepackage{ulem}

\journal{\dots}

\begin{document}

\begin{frontmatter}

\title{S-SIRUS: an explainability algorithm for spatial regression Random Forest}

\author[label1]{Luca Patelli}
\author[label2]{Natalia Golini \corref{cor1}}
\author[label2]{Rosaria Ignaccolo}
\author[label3]{Michela Cameletti}

\cortext[cor1]{natalia.golini@unito.it}

\affiliation[label1]{organization={Department of Economics and Management, University of Pavia},
            city={Pavia},
            country={Italy}}
\affiliation[label2]{organization={Department of Economics and Statistics ``Cognetti de Martiis", University of Torino},    city={Torino},
            country={Italy}}
\affiliation[label3]{organization={Department of Economics, University of Bergamo},
            city={Bergamo},
            country={Italy}}

\begin{abstract}
Random Forest (RF) is a widely used machine learning algorithm known for its flexibility, user-friendliness, and high predictive performance across various domains. However, it is non-interpretable.
This can limit its usefulness in applied sciences, where understanding the relationships between predictors and response variable is crucial from a decision-making perspective.
In the literature, several methods have been proposed to explain RF, but none of them addresses the challenge of explaining RF in the context of spatially dependent data. Therefore, this work aims to explain regression RF in the case of spatially dependent data by extracting a compact and simple list of rules. In this respect, we propose S-SIRUS, a spatial extension of SIRUS, the latter being a well-established regression rule algorithm able to extract a stable and short list of rules from the classical regression RF algorithm. A simulation study was conducted to evaluate the explainability capability of the proposed S-SIRUS, in comparison to SIRUS, by considering different levels of spatial dependence among the data. The results suggest that S-SIRUS exhibits a higher test predictive accuracy than SIRUS when spatial correlation is present. Moreover, for higher levels of spatial correlation, S-SIRUS produces a shorter list of rules, easing the explanation of the mechanism behind the predictions.
\end{abstract}


\end{frontmatter}

%\linenumbers

\section{Introduction}\label{sec:intro}
Researchers from different fields are moving towards the ever-increasing use of machine learning and deep learning models. These models exhibit high adaptability to data and can capture complex predictor-response relationships, potentially achieving high levels of accuracy \citep{sarker2021machine,sarker2021deep}. However, due to the difficulty in understanding how the prediction mechanism operates, they are defined as non-interpretable models or ``black boxes". Consequently, there has been a growing need to explain these models as comprehensively as possible, especially in critical domains such as healthcare, finance, precision agriculture, and others. It is important to note that there is no consensus in the literature regarding a rigorous definition of explainability. For an overview of the different principles, definitions and methods for explaining black box models see 
\cite{guidotti2018}, \cite{lipton2018}, \cite{murdoch2019}, \cite{burkart2021survey}, \cite{molnar2022} and \cite{rudin2022interpretable}.
Interestingly, our feeling is that there is a growing tendency to use black box models, even when well-established statistical models (interpretable by definition) provide equally accurate predictive results (see, for example, \citealt{ wikle2023agnostic}).

In this paper, we consider Random Forest (RF, \citealp{breiman2001random}), a machine learning algorithm known for being a very accurate predictive model but a black box. RF is an ensemble model that combines a large number of decision trees (say $B$). Each tree is grown by implementing recursive binary splitting (defining a partition of the predictor space) on a bootstrapped sample of the data. Each single split involves a predictor-cutpoint combination where the candidate predictor is chosen in a random subset of predictors.
When using RF for a regression problem, predictions for new observations are obtained by averaging the predictions obtained from the $B$ trees. As an ensemble of trees, RF achieves higher levels of accuracy compared to single trees, which are known to be characterized by high variability. The negative aspect is that, given the large number of trees that are aggregated to grow the forest, it is not feasible to inspect the structure of all the trees and directly understand the mechanism behind the predictions. 

The research community has already produced some contributions trying to explain RF and to understand how the available predictors contribute to the predictions. In this regard, in their review \cite{haddouchi2019survey} proposed a taxonomy of the methods that can be used to ``uncover insights in the RF resulting models". Moreover, \cite{ARIA2021100094} describe explainability strategies for RF and compare two rule extraction approaches: inTrees \citep{deng2019interpreting} and Node harvest \citep{meinshausen2010nodeharvest}.
More recently, another rule algorithm for classification and regression problems, namely SIRUS, has been proposed by \cite{benard21regression,benard2021SIRUS}.
However, an important issue has been left unexplored in all the approaches proposed for explaining RF: in environmental science applications, the phenomena to be studied may present spatial dependence, and the available data may be correlated in space. This contrasts with the classical RF algorithm (as defined by \citealp{breiman2001random}), which implicitly assumes that the data are independent. Therefore, it is necessary to seek a spatially oriented extension of RF for situations where observations are spatially dependent. In this regard, \cite{patelli2023path} propose a taxonomy for classifying all the strategies available in the literature for adapting the regression RF algorithm to the case of spatially correlated data. 
Among them, we adopt here the RF-GLS algorithm by \cite{saha2023random} that modifies internally the RF algorithm by moving from an ordinary least squares (OLS) version of RF to a generalized least squares (GLS) problem. This makes it possible to include in the fitting phase a variance-covariance matrix providing information about the spatial correlation.
However, a suitable strategy is necessary to explain this spatially aware black box.

This work aims to explain regression RF when space matters. In this regard, we consider the Stable and Interpretable RUle Set (SIRUS) algorithm for regression \citep{benard2021SIRUS}. SIRUS has been shown to achieve accuracy levels comparable to those observed in cutting-edge rule algorithms, such as Node harvest \citep{meinshausen2010nodeharvest} and RuleFit \citep{friedman2008predictive}, while generating more stable and compact rule lists.
In particular, our proposal is to combine SIRUS with RF-GLS in order to define an explainability strategy for RF which takes into account, in the fitting phase, the data spatial structure. Our proposal will be named S-SIRUS, which stands for Spatial SIRUS. Through a simulation study with pseudo-real data, we show that when the relative importance of the spatial structure varies, the obtained lists of rules are different, both in the size and structure, between SIRUS and S-SIRUS, also affecting the test predictive performance when the obtained set of rules is used for prediction.

The paper is structured as follows. In Section~\ref{sec:SpatialSIRUS} the proposal S-SIRUS algorithm for geostatistical data is presented.  Section~\ref{sec:simulation} contains the simulation setup and the results for the considered scenarios.
In Section~\ref{sec:conclusion} some concluding remarks are given.

%%%%%%%%%
\section{Spatial SIRUS}\label{sec:SpatialSIRUS}

In this work, we propose S-SIRUS to extract a stable and short list of rules from an RF-GLS, an RF for spatially correlated data. 
We focus on geostatistical data collected at $n$ spatial locations, $s_1, \ldots, s_n$, over a region $\mathcal{A} \subset \mathbb{R}^d$ (usually $d = 2$ and $s_i$ is the (\textit{latitude}$_i$, \textit{longitude}$_i$) vector). By denoting with $y(s_i)$ and $\mathbf{x}(s_i) = (x_{1}(s_i), \dots, x_{P}(s_i))^T$ the response value and 
the $P$-dimensional predictor vector for the $i$-th observation, we assume the following spatial model for the observed sample of data given by $\mathscr{D}_n = \{y(s_i),\mathbf{x}(s_i)\}_{i = 1}^{n}$:
\begin{equation}\label{Eq:mainmodel}
y(s_i) = f(\mathbf{x}(s_i)) + \omega^{\star}(s_i), \qquad i=1, \ldots, n,
\end{equation}
where $f(\cdot)$ is a function of the $P$ predictors (also known as large-scale component or regression function), and $\omega^{\star}$ is a spatially correlated process given by the sum of a spatial Gaussian process $\omega$, and a measurement error $\epsilon$ that is a Gaussian white noise.

With the proposed S-SIRUS, we aim to estimate the large-scale component using a small number of stable rules. S-SIRUS implements the same steps of SIRUS for regression problems \citep{benard21regression}, with the main novelty regarding the rule generation (see Step 1 in Section~\ref{subsec:ssirusfit}) where RF-GLS is adopted instead of the classical RF.

Before describing the steps of S-SIRUS to estimate the regression function, we introduce the concept of rule and briefly describe RF-GLS.
Rule algorithms are non-linear methods defined as a collection of elementary rules that are highly predictive 
and exhibit a simple structure. An elementary rule is a set of constraints on predictors, which forms a hyperrectangle in the predictor space where the associated prediction is constant. An elementary rule typically takes the form of an \textit{If-Then-Else} expression, where the \textit{If} condition involves at least one predictor. As SIRUS, S-SIRUS leverages the fact that each node (internal or terminal) of each tree of an RF can be converted into an elementary rule.

Being in a spatial framework, we consider RF-GLS that modifies substantially the RF algorithm to account for spatial correlation \citep{saha2023random}. RF-GLS is an extension of RF which makes use of GLS. The main idea is to represent the local (intra-node) loss as a global
(in all nodes) OLS linear regression problem with a binary design matrix. Then, the OLS loss is replaced by a GLS loss accounting for the spatial covariance, and the bootstrap resampling is applied to decorrelated response values. 
In an extensive simulation study, \cite{saha2023random} demonstrated that RF-GLS outperforms standard RF and other strategies adopted to adjust regression RF to spatially dependent data. This occurs both for the prediction of the large-scale component $f(\cdot)$ and of the response variable $y(\cdot)$, where the latter case requires the fitting of an ordinary kriging model to the residuals (see Section 2.6 in \citealp{saha2023random}).  
\subsection{Fitting of the rules via S-SIRUS} \label{subsec:ssirusfit}
We describe in the following the four steps of the S-SIRUS fitting algorithm whose implementation is summarized in Algorithm \ref{alg:sirusfit}.\\

\noindent \textbf{Step 1. Rule generation}: all the (internal or terminal) nodes from a large number $B$ of trees of an RF-GLS are extracted to generate a large collection of rules (commonly in the order of $10^4$). The choice of considering RF-GLS, instead of RF, represents our main proposal. For the definition of the rules, it is important to recall that each tree node in the forest defines a hyperrectangle in the predictor space $\mathbb{R}^P$.
Consequently, each node can be transformed into an elementary regression rule. Formally, each tree node is represented by a path, denoted by $\mathscr{P}$, enumerating the sequence of splits to reach the node from the tree's root. Here onward, we denote by $\Pi$ the finite list of all the possible paths, where each $\mathscr{P} \in \Pi$ defines a regression rule denoted by $\hat{g}_{n,\mathscr{P}}(\cdot)$.

As in SIRUS, the stability of the forest structure is enhanced by constraining node splits. For continuous predictors, this is achieved by dividing the range of each predictor into $q$ intervals using the empirical quantiles and replacing the predictor values with the upper limits of the intervals to which they belong. \cite{benard21regression} suggest to set $q=10$. No transformation is required instead for discrete and categorical predictors, as the set of possible split values is already finite. \\

\textbf{Step 2. Rule selection}: the most relevant regression rules are selected from the large collection generated in Step 1. Due to the quantile discretization, there is redundancy in the extracted rules. Frequently appearing rules indicate strong and persistent data patterns, making them ideal for forming a compact, stable, and predictive rule ensemble. Let $\hat{p}_{B,n}(\mathscr{P})$ be the relative occurrence frequency for each  $\mathscr{P} \in \Pi$, calculated on $B$ trees, and $p_0 \in (0,1)$ the frequency threshold. It follows that the set of the relevant rules is given by
\begin{equation*} 
    \hat{\mathscr{P}}_{B,n,p_{0}} = \lbrace \mathscr{P} \in \Pi : \hat{p}_{B,n}(\mathscr{P}) > p_0  \rbrace.
    \label{eq:frequentrules}
\end{equation*}
The threshold $p_0$ is a hyperparameter that needs fine-tuning (see Section \ref{subsec:ssiruscv}). Rules with more constraints (i.e., involving many splits) are more sensitive to data perturbation and thus appear less frequently in the trees. Therefore, optimal $p_0$ values typically select rules with one or two splits. Similarly to SIRUS, S-SIRUS considers shallow trees with a maximum depth of 2, thus reducing the computational cost while maintaining a nearly identical list of rules with respect to deeper trees. \\

\textbf{Step 3. Rule set post-treatment}: rules in $\hat{\mathscr{P}}_{B,n,p_{0}}$ whose corresponding hyper-rectangles overlap are linearly dependent and are filtered. Otherwise said, the path $\mathscr{P}$ is removed from $\hat{\mathscr{P}}_{B,n,p_{0}}$ if the associated rule is a linear combination of rules related to paths with a higher occurrence frequency. \\

\textbf{Step 4. Rule aggregation}: the resulting filtered rules are linearly combined to obtain a single estimate of the large-scale component $f(\cdot)$. In particular, 
the final aggregated estimate of $f(\cdot)$ is given by
\begin{equation}
    \hat{f}_{B,n,p_0}(\cdot) = \hat{\beta}_0 + \Sigma_{\mathscr{P} \in \hat{\mathscr{P}}_{B,n,p_{0}}} \hat{\beta}_{n,\mathscr{P}}\cdot\hat{g}_{n,\mathscr{P}}(\cdot),
    \label{eq:ridgeregression}
\end{equation}
\noindent where $\hat{\beta}_0$ and $\hat{\beta}_{n,\mathscr{P}}$ are the coefficients of the linear combination of rules estimated by ridge regression and constrained to have non-negative values to enhance the explainability of the algorithm, and the rules $\hat{g}_{n,\mathscr{P}}(\cdot)$ play the role of predictors. Let us recall that a rule $\hat{g}_{n,\mathscr{P}}(\cdot)$, associated with a path $\mathscr{P}$, is a piecewise constant function whose value depends on which hyper-rectangle $H_{\mathscr{P}} \subset \mathbb{R}^{P}$ a generic predictor vector $\mathbf{x}$ falls into.
If $\mathbf{x} \in H_{\mathscr{P}}$, then the rule returns the average of the $y(s_i)$'s associated to the $\mathbf{x}(s_i)$'s falling into $H_{\mathscr{P}}$; otherwise, the rule returns the average of the $y(s_i)$'s associated to the $\mathbf{x}(s_i)$'s that do not fall into $H_{\mathscr{P}}$. The estimation via ridge regression allows a stable aggregation of rules while the sparsity of the solution of S-SIRUS  is controlled by $p_0$.\\

Being inspired by SIRUS, S-SIRUS satisfies the three minimum requirements for explaining a black box, here given by RF-GLS: simplicity, stability and predictivity \citep{bin2020,benard2021SIRUS,benard21regression}. Firstly, S-SIRUS is simple as it generates a small number of rules, while simplicity is explicitly regulated by the hyperparameter $p_0$. Secondly, S-SIRUS is asymptotically stable with respect to data perturbation: when the sample size is large enough, it provides the same list of rules across several fits on independent samples (see Theorem 1 in \citealp{benard21regression}). In practical scenarios, obtaining additional independent samples is rare and, therefore, a $K$-fold cross-validation (usually with $K=10$) is employed to simulate data perturbation. The stability metric ($stability.metric$ in Algorithm \ref{alg:sirusfit}) in S-SIRUS is defined empirically as the average proportion of rules shared by two S-SIRUS fits across different CV folds. This metric, known as the Dice-Sorensen index, equals 1 when the same set of rules is selected across all the $K$ folds and 0 when there is no overlap in rules between any two folds. As in SIRUS, the S-SIRUS stability increases with the number of trees. The same stopping criterion designed for SIRUS is then applied, aiming to grow the minimum number of trees necessary to ensure that, on average, $(1 -\alpha)\%$ of the rules are shared between two S-SIRUS fits (the recommended value for $\alpha$ is $0.05$); refer to Section 4 of \cite{benard2021SIRUS} for more details.
Finally, as in SIRUS, the predictive performance is measured as the proportion of unexplained variance, which serves as a natural measure of prediction error. It is important to note that the stopping criterion based on stability also maximizes the predictive performance of S-SIRUS. This is because the number of trees required to reach the desired level of stability is greater than the number needed to obtain a high level of predictive performance (see Section 7 of the Supplementary Material of \citealp{benard21regression}).\\

Algorithm~\ref{alg:sirusfit} outlines the four steps of the S-SIRUS fitting procedure in the case when only continuous predictors are available. It is implemented using the \texttt{R} function \texttt{s.sirus.fit}, which works with trees of maximum depth equal to 2 and can also handle both discrete and categorical predictors. The \texttt{s.sirus.fit} function is available in the GitHub repository at \url{https://github.com/LucaPate/Spatial_SIRUS}. This repository contains all the \texttt{R} functions required to implement S-SIRUS. They are a modified version of the functions available in the \texttt{sirus R} package (e.g., \texttt{s.sirus.fit}) to enable the use of RF-GLS instead of RF. Specifically, RF-GLS is performed using the \texttt{RFGLS\_estimate\_spatial} function from the \texttt{RandomForestsGLS R} package \citep{saha2022rpackage}.

\begin{algorithm}[ht!]
\footnotesize{
\caption{S-SIRUS fitting}\label{alg:sirusfit}
\hspace*{\algorithmicindent} \textbf{Input:} \\
\hspace*{\algorithmicindent} $-~\{s_i\}_{i = 1}^{n}$: coordinates of the spatial locations \\
\hspace*{\algorithmicindent}  $-~\mathscr{D}_n = \{y(s_i),\mathbf{x}(s_i)\}_{i = 1}^{n}$: set of 
spatially correlated data \\
\hspace*{\algorithmicindent} $-~q$: number of intervals for the predictor discretization \\
\hspace*{\algorithmicindent} $-~B$: number of trees grown in the forest \\
\hspace*{\algorithmicindent} $-~b$: number of trees grown between two evaluations of the stopping criterion \\
\hspace*{\algorithmicindent} $-~p_0$: occurrence frequency threshold\\
\hspace*{\algorithmicindent} \textbf{Procedure:}
\begin{algorithmic}[1]
\State Create \textit{data.bin.y} from $\mathscr{D}_n$ as the result of the  discretization of the $P$ continuous predictors in $q$ values
\If{$B = NULL$}
\State $stability.metric$ = 0
\State $B = 0$
\While{$stability.metric < (1 -\alpha)\%$}
\State Grow $b$ trees of an RF-GLS giving as input \textit{data.bin.y} and $\{s_i\}_{i = 1}^{n}$
\State Extract all nodes from the $b$ trees and define all the paths  $\mathscr{P}$ in $\Pi_b$
\State Calculate the absolute occurrence frequency for each $\mathscr{P} \in \Pi_b$ 
\State $B \leftarrow B + b$
\If{$B = b$}
\State Calculate $stability.metric$ based on $B$ trees and the absolute occurrence frequency  of each $\mathscr{P} \in \Pi_B$ 
\Else
\State Merge the sets of paths: $\Pi_B = \Pi_{B-b} \cup \Pi_b$
\State Calculate $stability.metric$ based on $B$ trees and the absolute occurrence frequency  of each $\mathscr{P} \in \Pi_B$ 
\EndIf
\EndWhile
\Else 
\State Grow $B$ trees of an RF-GLS giving as input \textit{data.bin.y} and $\{s_i\}_{i = 1}^{n}$
\State Extract all the nodes from the $B$ trees, define each path $\mathscr{P}$ in $\Pi_B$
\State Calculate the absolute occurrence frequency $\forall$ $\mathscr{P} \in \Pi_B$ 
\EndIf
\State Calculate $\hat{p}_{B,n}(\mathscr{P})$ and use them to sort $\Pi_B$  
\State  Extract the most frequent rules according to $p_0$ defining $\hat{\mathscr{P}}_{B,n,p_0}$ 
\State  Filter $\hat{\mathscr{P}}_{B,n,p_0}$ for removing the linearly dependent rules 
\State  Perform ridge regression (see Eq.(\ref{eq:ridgeregression}))
\end{algorithmic}
\hspace*{\algorithmicindent} \textbf{Output:} 
List of rules (\textit{If-Then-Else} structure); $\hat{p}_{B,n}(\mathscr{P})$; $\hat \beta_0$ and $\hat \beta_{n,\mathcal{P}}$}
\end{algorithm}

%%%%%%%%%
\subsection{S-SIRUS cross-validation for $p_0$} \label{subsec:ssiruscv}
As anticipated in the previous section, the threshold $p_0 \in (0,1)$ is a hyperparameter of S-SIRUS that needs to be fine-tuned. This threshold controls the sparsity of the solution of S-SIRUS by selecting a compact set of the relevant rules in the forest, $\hat{\mathscr{P}}_{B,n,p_0}$. To ensure that the resulting list of rules remains interpretable, the range of possible values for $p_0$ is set so that the maximum number of rules does not exceed $25$. Then, the optimal value of $p_0$ is obtained from a fine grid of values by a $K$-fold cross-validation using a standard bi-objective optimization procedure to maximize both stability and predictivity and choosing $p_0$ to be as close as possible to the ideal case of 0 unexplained variance and $90\%$ stability (see \citealp{benard21regression} for more details). To achieve a robust assessment of $p_0$, the CV is repeated more times (e.g., 10 times), and the median $p_0$ value across these repetitions is considered. This S-SIRUS CV for tuning the hyperparameter $p_0$ is implemented using the \texttt{R} function \texttt{s.sirus.cv} available in our GitHub repository.

When speaking about CV for spatially correlated data a note about the sampling strategy is necessary. In S-SIRUS we adopt a standard CV sampling as in the case of independent data, following the approach suggested by \cite{rabinowicz2022cross}. In fact, they show that standard CV is unbiased, even in the case of correlated data, when the CV partitioning maintains the distributional relation between the prediction (test) set and the training set. In spatial applications, this condition is satisfied when the training and test observations are randomly sampled from the same region where a realization of the spatially correlated process $\omega^{*}$ occurred. 

\subsection{Response variable prediction with S-SIRUS} \label{subsec:ssiruspredict}
The prediction of the large-scale component for a new observation, i.e., at an unmonitored location with coordinates $s_0$ and predictor vector $\mathbf{x}(s_0)$, is straightforward with S-SIRUS by using Eq.~\eqref{eq:ridgeregression}. However, in spatial applications, the interest is often in predicting the response variable $y(s_0)$. In this case, it is necessary to adopt a post-processing strategy as described in \cite{patelli2023path}.
Indeed, first, the large-scale component is predicted by using S-SIRUS, then an ordinary kriging model is fitted over the S-SIRUS residuals. The final prediction for the response variable $y$ for a new observation is then given by
\begin{equation}\label{Eq:mainmodel_prediction}
\hat{y}(s_0) = \hat{f}_{B,n,p_0}(\mathbf{x}(s_0)) + \hat{\omega}^{\star}(s_0),
\end{equation}
\noindent where $\hat{f}_{B,n,p_0}(\mathbf{x}(s_0))$ is obtained by applying Eq.~\eqref{eq:ridgeregression} and $\hat{\omega}^{\star}(s_0)$
is the kriged residual after fitting an ordinary kriging model.

S-SIRUS large-scale component predictions for new observations
can be computed using the \texttt{s.sirus.predict R} function available in our GitHub repository.
To obtain a value for $\hat{\omega}^{\star}(s_0)$ one can use, for example, the \texttt{R} spatial regression functions in the packages \texttt{gstat} \citep{pebesma2004multivariable} or \texttt{BRISC} \citep{saha2018brisc}.

%%%%%%%%%
\section{Simulation study}
\label{sec:simulation}
To evaluate the performance of the spatially aware S-SIRUS algorithm compared to the original SIRUS, we propose a simulation study using pseudo-real data that mimics a real phenomenon with different levels of spatial correlation.

\subsection{Data generation}
The AgrImOnIA dataset\footnote{The AgrImOnIA dataset is openly available on Zenodo platform (\url{https://doi.org/10.5281/zenodo.7956006}).}\citep{agrimoniadatapaper2023}  provides data about livestock, meteorology and air quality in the Lombardy region (Italy). Daily data are available from 2016 to 2021 for 141 monitoring stations placed in the region and a neighbouring buffer zone. Among the available data, we consider the daily mean concentration of Particulate Matter with an aerodynamic diameter less than 10 $\mu g$ (PM$_{10}$, response variable), the location of the monitoring station (latitude and longitude, and altitude) and the main meteorological variables generally used in air pollution modeling (see e.g., \citealt{FIORAVANTI2021, otto2024spatiotemporal}). Table~\ref{tab:regressors} provides a description of all the variables used in the simulation. Being interested in purely spatial analysis, since winter is the critical season for PM$_{10}$ concentrations and to have the largest sample size in a day, we selected data recorded on 15-02-2019.
\begin{table}[!ht]
\caption{Description of the variables from the AgrImOnIA dataset, available on the selected day 15-02-2019, and used for the simulation study.}
    \centering
    \scriptsize{
    \begin{tabular}{l|l}
    \toprule
    Variable short name & Variable description [unit of measurement] \\
    \midrule
    \texttt{PM$_{10}$} & Particulate matter (aerodynamic diameter $<$ 10 $\mu g$) concentration [$\mu g/m^3$]\\
    \texttt{Latitude} & North-south position [$DD$]\\
    \texttt{Longitude} & East-west position [$DD$]\\
    \texttt{Altitude}   &  Height in relation to sea level [m] \\
    \texttt{blh\_layer\_max}   & Daily maximum height of the air mixing layer [m] \\
    \texttt{temp\_2m}   &  Air temperature at 2 m [\textit{°C}] \\
    \texttt{rh\_mean}   & Relative humidity [\textit{\%}] \\
    \multirow{2}{*}{\texttt{solar\_radiation} }  & Amount of solar radiation that reaches a horizontal plane at the surface \\
    & minus the amount reflected by the Earth’s surface [$J/m^2$] \\
    \texttt{surface\_pressure}   & Pressure (force per unit area) of the atmosphere at the surface of land [\textit{Pa}] \\
    \texttt{wind\_speed\_100m\_mean}   &  Average wind speed at 100 m [$m/s$] \\
    \texttt{nox\_sum}   &  Emissions of NO$_x$ across all sectors (anthropogenic) [$mg/m^2$] \\
    \bottomrule
    \multicolumn{2}{l}{$\mu g/m^3$ = microgram per cubic meter; $DD$ = decimal degrees; $m$ = meters; \textit{°C} = Celsius degrees; } \\
    \multicolumn{2}{l}{$\%$ = percentage; $J/m^2$ = joules per square meter; \textit{Pa} = Pascal; $m/s$ = meter per second;  } \\
    \multicolumn{2}{l}{$mg/m^2$ = milligram per square meter.}
    \end{tabular}
    }
    \label{tab:regressors}
\end{table}
All the AgrImOnIA data, apart from PM$_{10}$ concentrations, are also available for a regular grid composed of 1053 locations. Among them, we randomly select $n=500$ sites where to simulate $\log$(PM$_{10}$) values consistently with Eq.~\eqref{Eq:mainmodel}. As far as it concerns the large-scale component, $f(\mathbf{\cdot})$, we first fit a Generalized Additive Model (GAM) to the data available on the selected day gathered at $106$ locations (i.e., monitoring stations where the response variable is measured). In the GAM model, the response variable is the log-transformed PM$_{10}$, and the predictors are the (standardized) variables reported in Table~\ref{tab:regressors}, coordinates excluded. By performing a backward selection, we include in the resulting model only three variables, namely air temperature at $2$ m, surface pressure and altitude, the latter considered with a non-linear relationship with the response variable. Fitting results are obtained by the \texttt{R} function \texttt{gam} from the \texttt{mgcv package} and are shown in \ref{appendix:gam}.

The GAM estimates are then used for simulating the large-scale values at the considered locations ($i=1, \ldots, 500)$ according to the following equation: 
\begin{equation} \label{eq:gam}
f^{sim}(\mathbf{x}(s_i)) = \beta_0 + \beta_1\texttt{temp\_2m}(s_i) + \beta_2\texttt{surface\_pressure}(s_i) +  g(\texttt{Altitude}(s_i)),
\end{equation}
where $\beta_0=4.02051$, $\beta_1 = -0.35528$, $\beta_2= 0.59869$ and $g(\cdot)$ is a (fitted) smoothing spline with 6.651 effective degrees of freedom. Then, the value of the response variable for location $s_i$ is generated as follows:
\begin{equation}\label{eq:sim}
y^{sim}(s_i) = f^{sim}(\mathbf{x}(s_i)) + \omega^{sim}(s_i) + \epsilon^{sim}(s_i), 
\end{equation}
where $\omega^{sim}(s_i)$ is simulated from a spatial Gaussian process with exponential covariance function \citep{cressie1993statistics}, variance $\sigma^2_\omega$ and spatial range parameter $\phi$; $\epsilon^{sim}(s_i)$ is drawn from a zero-mean i.i.d. Gaussian random noise with variance $\sigma^2_\epsilon$. 

To assess the impact of spatial correlation on rules extraction, three scenarios (A, B and C) are considered with different values of the Signal-to-Noise-ratio (SNR) defined as 
\begin{equation}\label{eq:SNR}
    \text{SNR} = \frac{ Var(f(\mathbf{X}))}{\sigma^2_\omega},
\end{equation}
\noindent where $\mathbf{X}$ is the design matrix for the $n$ observations, $Var(f(\mathbf{X}))$ is the large-scale variance and $\sigma^2_\omega$ the spatial process variance. SNR measures the relative importance of the large-scale component compared to the spatial one, as also discussed in the simulation study by \cite{saha2023random}. We consider three values of SNR: in Scenario A the large-scale variability is half of the spatial variance, in Scenario B they are equal, while in Scenario C the large-scale variability is twice the spatial variance. We expect S-SIRUS to outperform SIRUS in Scenario A when the spatial term contributes the most to the variability of the response variable.

Table~\ref{tab:scenario} shows the parameter values and related SNR for the three scenarios. To obtain the three SNRs, we obtain $Var(f(\mathbf{X}))$ by computing the variance of the 500 $f^{sim}(\mathbf{x}(s_i))$ values, which turns out to be equal to $0.0864$. By fixing SNR equal to $1/2,1$ and $2$, we then derive the values for $\sigma^2_\omega$ from Eq.~(\ref{eq:SNR}). Moreover, as in \cite{saha2023random}, we set the measurement error variance $\sigma^2_\epsilon$ equal to 10\% of $\sigma^2_\omega$. Finally, the spatial range parameter $\phi$ is set to $1/50$; this corresponds to a practical range, where the correlation is near $0.05$, of $150$ km, with the maximum distance in the considered spatial domain being $413.6$ km.

\begin{figure}[!h]
\caption{Spatial sites coloured according to the values of the simulated response variable ($\log(\text{PM}_{10})$) for Scenario A; 400 training observations (circle) and 100 test observations (triangle).}
    \centering
    \includegraphics[width=0.65\textwidth]{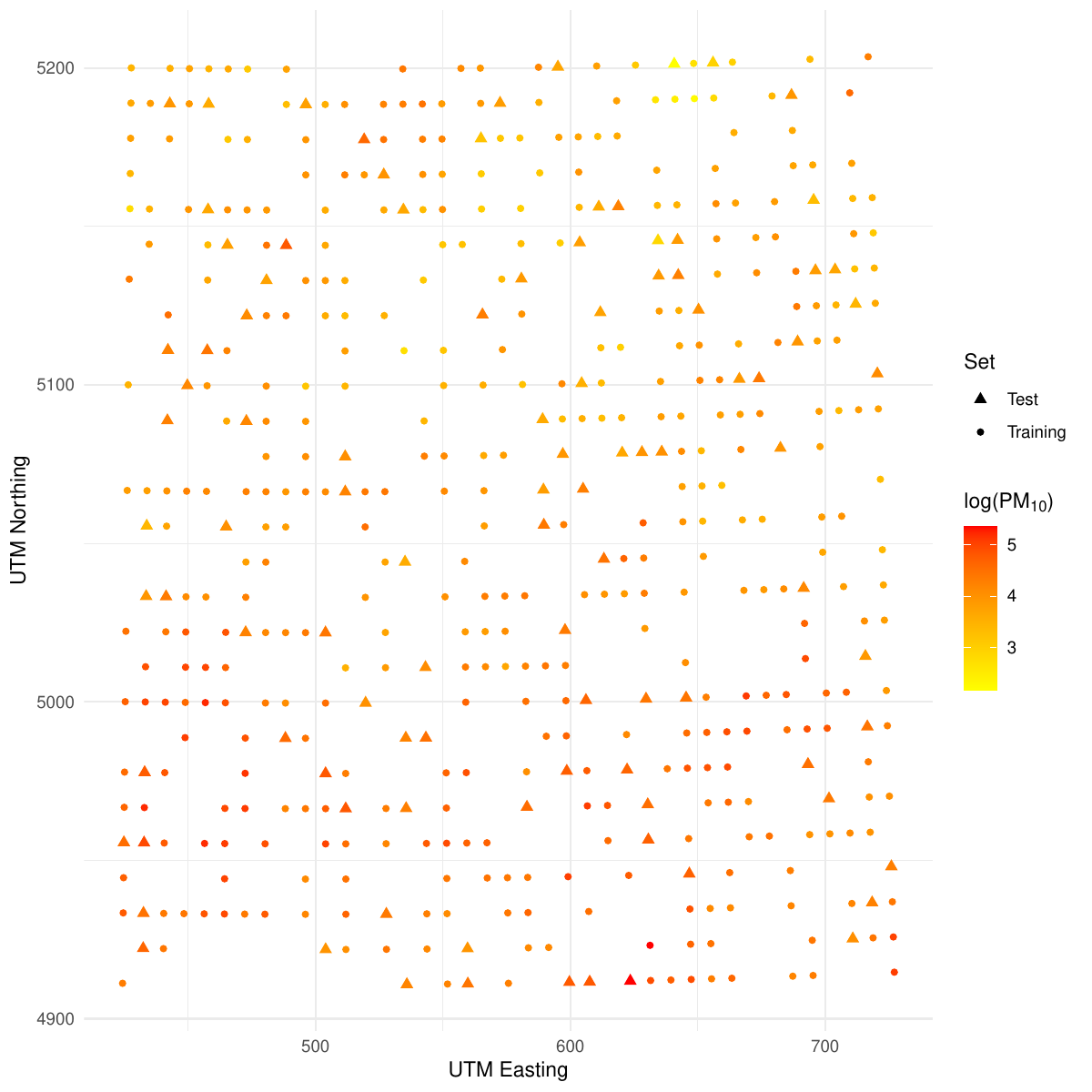}
    \label{fig:scenarioamap}
\end{figure}

\begin{table}[!ht]
\caption{Parameter values used in the simulation study for the three scenarios.}
\begin{center}
\begin{tabular}{c@{\quad}c@{\quad}c@{\quad}c@{\quad}c}
\toprule
Scenario & SNR & $\phi$ & $\sigma^2_{\omega}$ & $\sigma^2_{\epsilon}$ \\
\midrule
 A & 1/2 &  1/50 & 0.1728 & 0.0173 \\
 B & 1   &  1/50 & 0.0864 & 0.0086 \\
 C & 2   &  1/50 & 0.0432 & 0.0043 \\ %
\bottomrule
\end{tabular}
\end{center}
\label{tab:scenario}
\end{table}
For each scenario, we implement SIRUS and S-SIRUS considering all the (standardized) predictors reported in Table~\ref{tab:regressors}, and longitude and latitude are converted according to the Universal Transverse Mercator (UTM) system. 
We consider a training set of 400 observations randomly selected from the 500 simulated ($\mathscr{D}_{400}$), and use the remaining 100 observations for predictive purposes as a test set ($\mathscr{D}_{100}$). Figure~\ref{fig:scenarioamap} shows the training and test spatial sites coloured according to the values of $\log(\text{PM}_{10})$ in the Scenario A.

\newpage

\subsection{SIRUS and S-SIRUS results}\label{subsec:results}
In this section, we present the results of SIRUS and S-SIRUS applied to the three scenarios. 

\begin{figure}[!ht]
\caption{Unexplained variance (top panel) and stability (bottom panel) versus the number of rules for SIRUS (left panels) and S-SIRUS (right panels) in Scenario A for a fine grid of $p_0$, assessed via standard CV with $K=10$ folds. The results are averaged and bars show the variability of the metrics across 10 repetitions. 
The optimal $p_0$ value is the median $p_0$ value across the 10 CV repetitions (0.0257 for SIRUS and 0.0296 for S-SIRUS).}
      \centering
	   \begin{subfigure}{0.495\linewidth}	\includegraphics[width=\linewidth]{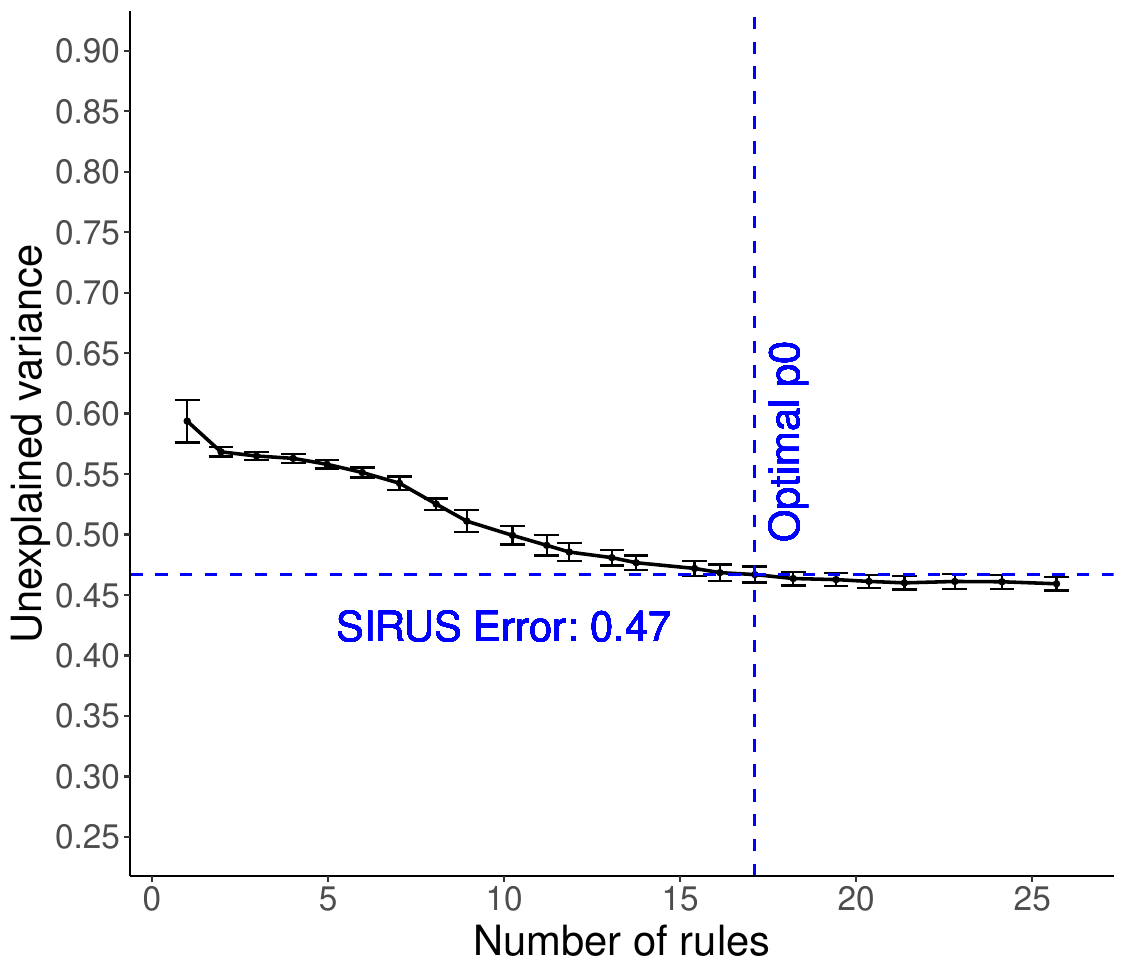}
		\caption{Unexplained variance SIRUS}
	   \end{subfigure}
	   \begin{subfigure}{0.495\linewidth}
		\includegraphics[width=\linewidth]{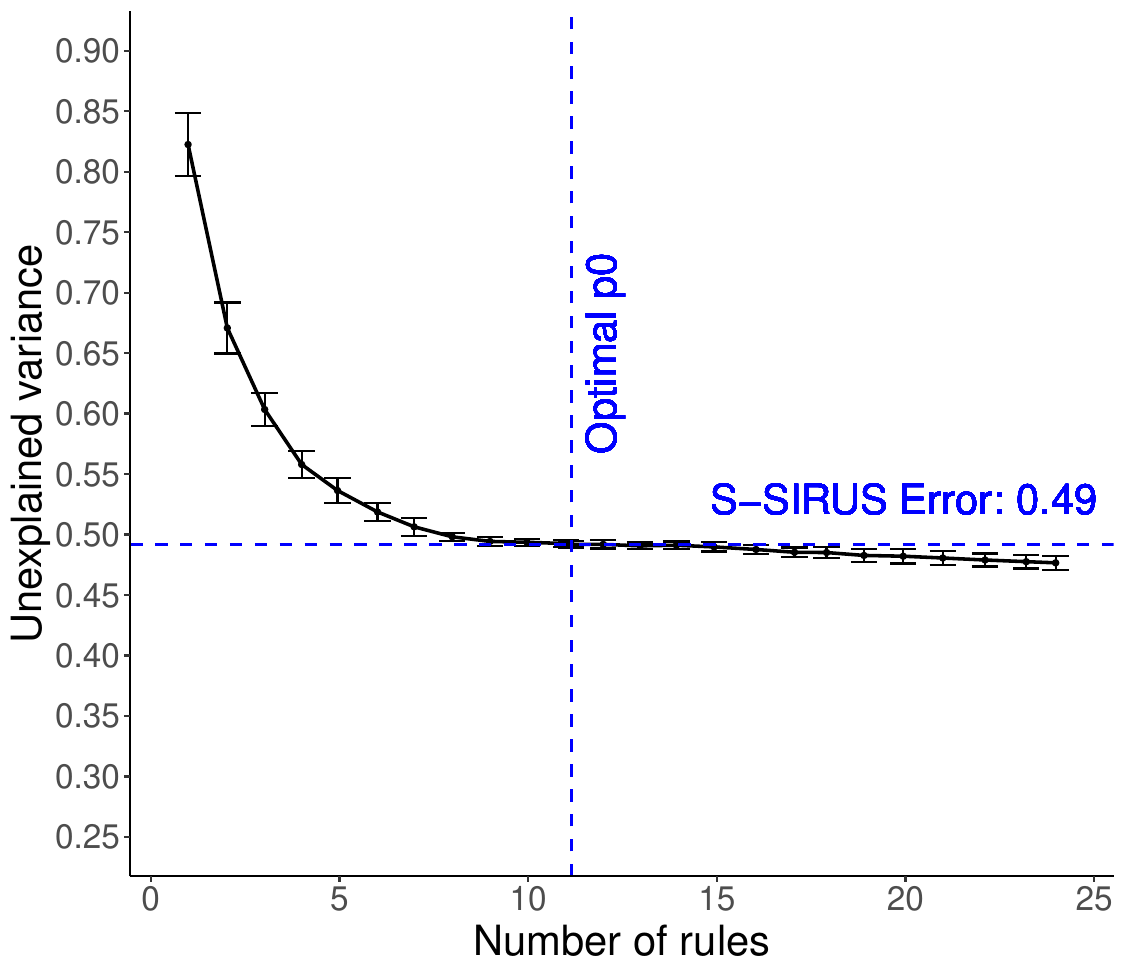}
		\caption{Unexplained  variance S-SIRUS}
	    \end{subfigure}
	\vfill
	     \begin{subfigure}{0.495\linewidth}
		 \includegraphics[width=\linewidth]{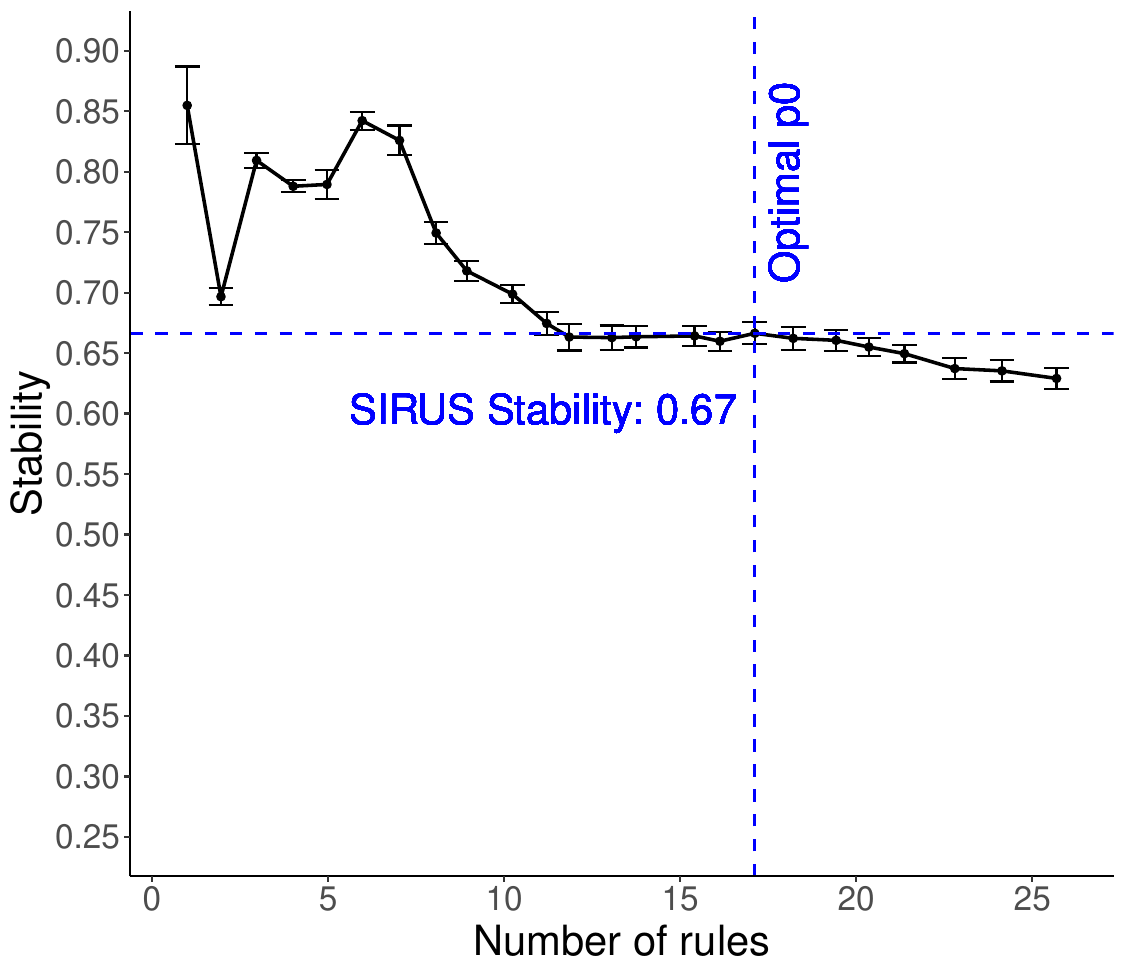}
		 \caption{Stability SIRUS}
	      \end{subfigure}
	       \begin{subfigure}{0.495\linewidth}
		  \includegraphics[width=\linewidth]{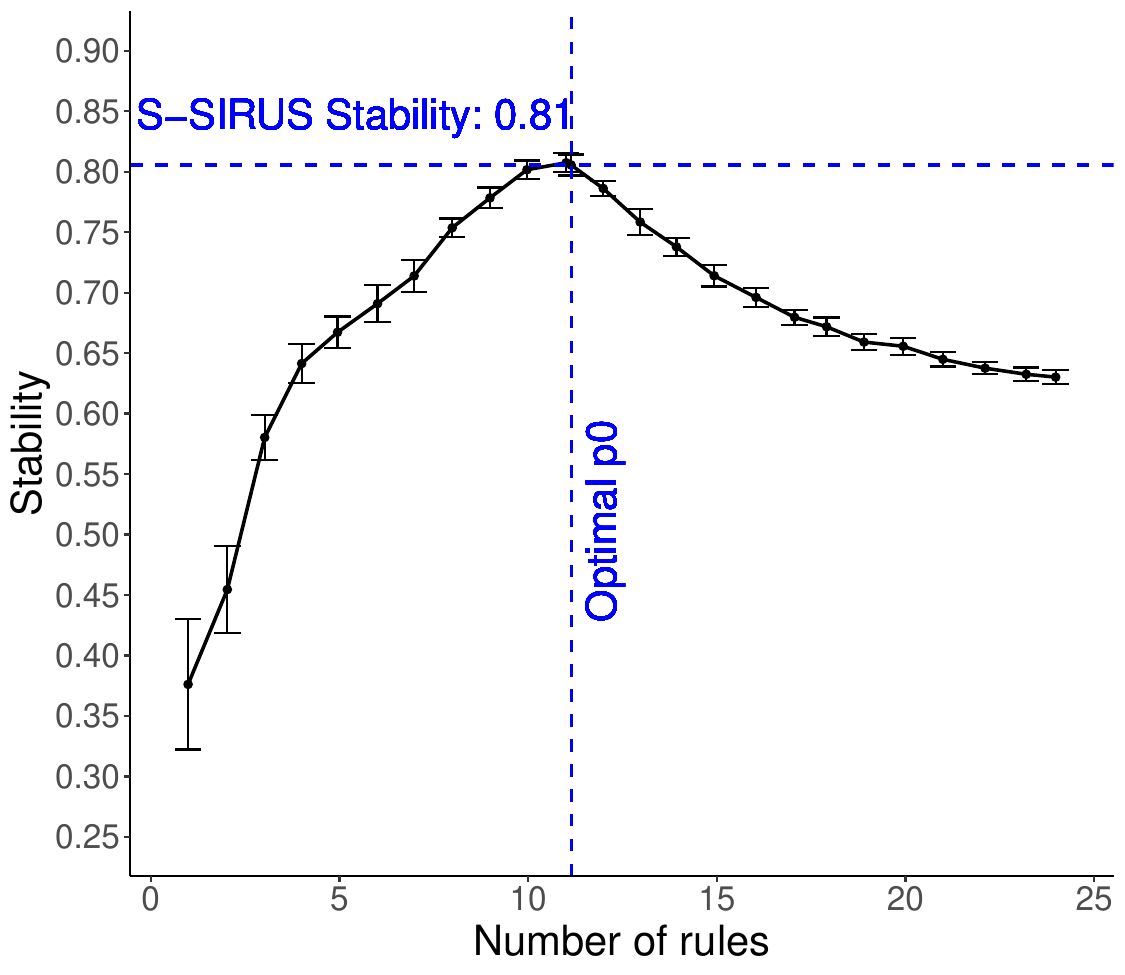}
		  \caption{Stability S-SIRUS}
	       \end{subfigure}
	\label{fig:cvfunctionoutA}
\end{figure}

Figure~\ref{fig:cvfunctionoutA} shows CV output for Scenario A with the predictive performance (top), expressed as unexplained variance, and stability (bottom) versus the number of rules of SIRUS (left panels) and S-SIRUS (right panels) as $p_0$ varies along a fine grid. These results are obtained by performing 10 repetitions of a standard 10-fold cross-validation on $\mathscr{D}_{400}$, as described in Section \ref{subsec:ssiruscv}. The optimal $p_0$ value is the median $p_0$ value across the 10 repetitions. According to predictivity and stability, we obtain two distinct optimal $p_0$ values resulting in different numbers of rules and levels of stability while maintaining similar unexplained variance values. In particular, even if SIRUS and S-SIRUS show a similar predictive error value (0.47 and 0.49, respectively), S-SIRUS needs a lower number of rules (11) to reach a higher stability (81\%). Similar results are obtained for Scenario B, where SNR = 1 (see Figure~\ref{fig:cvfunctionoutB}). Indeed, S-SIRUS still outperforms SIRUS in terms of stability while maintaining the same error level achieved with fewer rules. We observe an improvement in error levels (i.e., lower unexplained variance) for both algorithms. Regarding stability, SIRUS shows a slight improvement, whereas S-SIRUS experiences a slight reduction. As expected, in Scenario C, characterized by weaker spatial dependence (SNR = 2), SIRUS outperforms S-SIRUS, with respect to stability and simplicity (see Figure~\ref{fig:cvfunctionoutC}). 

Table~\ref{fig:cvfunctionoutA} exhibits the lists of rules returned by SIRUS (top panel) and S-SIRUS (bottom panel) for Scenario A, obtained considering $\mathscr{D}_{400}$, $q = 10$ and the optimal $p_0$ value equal to 0.0257 for SIRUS and 0.0296 for S-SIRUS. To reach convergence, $7000$ trees are grown for SIRUS and 28000 for S-SIRUS. 
To grow the SIRUS and S-SIRUS forests we consider a number of predictors equal to $P/3$, and $b = 1000$ trees between two evaluations of the stopping criterion.
In particular, the ``Intercept" and the column ``Weight" in Table~\ref{fig:cvfunctionoutA} refer to $\hat{\beta_0}$ and $\hat{\beta}_{B,\mathscr{P}}$ respectively, i.e., the coefficients of the linear combination of rules
estimated by ridge regression (see Eq.~\eqref{eq:ridgeregression}). The columns ``SIRUS Rule" and ``S-SIRUS Rule" provide the \textit{If-Then-Else} structure of each rule and the number of observations in $\mathscr{D}_{400}$ that either satisfy or not the rule's conditions. The rules are combined with the weights to obtain a single estimate of the large-scale component. Finally, the column ``Frequency" refers to $\hat{p}_{B,400}(\mathscr{P})$, i.e., the relative occurrence frequency of each rule calculated on $B$ trees. The SIRUS and S-SIRUS lists of rules share four of them (highlighted in grey), still having different weights and frequencies. Each rule explains how the predictors (weather conditions, NO$_{x}$ emissions and altitude) impact on the response variable ($\log$PM$_{10}$). It is notable that higher values of particulate matter concentrations are expected for increased levels of solar radiation, relative humidity, surface pressure, temperature and daily maximum height of the air mixing layer. Instead, lower values are expected at locations with higher altitudes. Additionally, SIRUS rules are more complex, with nine rules out of fifteen considering two constraints. Moving to Scenario B, where SNR = 1, the results are comparable (see Table~\ref{tab:rulesscenarioB}). Instead, the situation is the opposite in Scenario C, characterized by weaker spatial dependence (SNR = 2) with SIRUS returning a smaller number of rules involving only one predictor (see Table~\ref{tab:rulesscenarioC}). 

\begin{landscape}
\begin{table}[!ht]
    \centering
    \caption{SIRUS (top) and S-SIRUS (bottom) list of rules for Scenario A.}
    \scriptsize{
\begin{tabular}[t]{llc}
\toprule
\multicolumn{1}{c}{\textbf{Average}  $\log$(PM$_{10}$) =  4.08} & \multicolumn{1}{c}{\textbf{Intercept} = -3.71} & \textbf{n} = 400\\
\\
\multicolumn{1}{l}{\textbf{Weight}} & \multicolumn{1}{c}{\textbf{SIRUS Rule}} & \multicolumn{1}{c}{\textbf{Frequency}} \\
\midrule
\rowcolor{gray!20}
0.195 & if solar\_radiation $<$ 0.66 then 3.72 ($n=200$) else 4.44 ($n=200$) &0.246\\
\rowcolor{gray!20}
0.0572 & if rh\_mean $<$ -0.275 then 3.68 ($n=160$) else 4.34 ($n=240$)&0.135\\
0.091 & if surface\_pressure $<$ 0.286 then 3.75 ($n=200$) else 4.41 ($n=200$)&0.120\\
\rowcolor{gray!20}
0.0195 & if surface\_pressure $<$ -0.122 then 3.67 ($n=160$) else 4.35 ($n=240$)&0.089\\
\rowcolor{gray!20}
0.0917 & if blh\_layer\_max $<$ -0.173 then 3.62 ($n=120$) else 4.28 ($n=280$)&0.081\\
0.176 & if solar\_radiation $<$ 0.66 \& surface\_pressure $<$ -1.46 then 3.35 ($n=40$) else 4.16 ($n=360$) &0.060\\
0.0111 & if Altitude $<$ -0.332 then 4.39 ($n=200$) else 3.77 ($n=200$) &0.055\\
0.284 & if temp\_2m $\ge$ 0.329 \& wind\_speed\_100m\_mean $\ge$ -0.306 then 4.47 ($n=189$) else 3.73 ($n=211$) &0.046\\
0.152 & if solar\_radiation $\ge$ 0.66 \& nox\_sum $<$ 0.0438 then 4.51 ($n=140$) else 3.85 ($n=260$) &0.043\\
0.163 & if blh\_layer\_max $<$ 0.18 \& solar\_radiation $\ge$ 0.66 then 4.72 ($n=18$) else 4.05 ($n=382$) &0.037\\
0.0226 & if rh\_mean $\ge$ -0.275 \& wind\_speed\_100m\_mean $<$ -0.306 then 3.94 ($n=54$) else 4.1 ($n=346$) &0.032\\
0.267 & if solar\_radiation $<$ 0.66 \& nox\_sum $<$ -0.444 then 3.36 ($n=38$) else 4.15 ($n=362$) &0.031\\
0.175 & if temp\_2m $<$ 0.329 \& surface\_pressure $<$ -1.46 then 3.35 ($n=40$) else 4.16 ($n=360$) &0.030\\
0.167 & if surface\_pressure $\ge$ 0.286 \& nox\_sum $<$ 0.0438 then 4.52 ($n=131$) else 3.87 ($n=269$) &0.027\\
0.035 & if blh\_layer\_max $\ge$ -0.173 \& solar\_radiation $<$ 0.66 then 3.87 ($n=80$) else 4.13 ($n=320$) &0.026\\
\bottomrule
& &  \\
\end{tabular}
\begin{tabular}[t]{llc}
\toprule
\multicolumn{1}{c}{\textbf{Average}  $\log$(PM$_{10}$) =  4.08} & \multicolumn{1}{c}{\textbf{Intercept} = -2.76} & \textbf{n} = 400\\
\\
\multicolumn{1}{l}{\textbf{Weight}} & \multicolumn{1}{c}{\textbf{S-SIRUS Rule}} & \multicolumn{1}{c}{\textbf{Frequency}} \\
\midrule
0.171 & if nox\_sum $<$ -0.444 then 3.41 ($n=40$) else 4.15 ($n=360$)&0.119\\
\rowcolor{gray!20}
0.0475 & if blh\_layer\_max $<$ -0.173 then 3.62 ($n=120$) else 4.28 ($n=280$)&0.098\\
0.287 & if Altitude $<$ 1.53 then 4.17 ($n=360$) else 3.28 ($n=40$)&0.086\\
\rowcolor{gray!20}
0.11 & if surface\_pressure $<$ -0.122 then 3.67 ($n=160$) else 4.35 ($n=240$)&0.084\\
\rowcolor{gray!20}
0.418 & if solar\_radiation $<$ 0.66 then 3.72 ($n=200$) else 4.44 ($n=200$)&0.072\\
0.152 & if rh\_mean $<$ -1.04 then 3.52 ($n=80$) else 4.22 ($n=320$)&0.063\\
0.21 & if wind\_speed\_100m\_mean $<$ -0.306 then 3.79 ($n=160$) else 4.27 ($n=240$)&0.063\\
0.0452 & if surface\_pressure $<$ -1.11 then 3.53 ($n=80$) else 4.22 ($n=320$)&0.055\\
0.0193 & if temp\_2m $<$ 0.631 then 3.8 ($n=200$) else 4.36 ($n=200$)&0.048\\
\rowcolor{gray!20}
0.111 & if rh\_mean $<$ -0.275 then 3.68 ($n=160$) else 4.34 ($n=240$)&0.046\\
0.104 & if Altitude $<$ 1.17 then 4.22 ($n=320$) else 3.51 ($n=80$)&0.033\\
\bottomrule
\end{tabular}
}\label{tab:rulesscenarioA}
\end{table}
\end{landscape}

\begin{figure}
\caption{Estimated exponential variograms $\hat \gamma (h)$ for SIRUS (black solid line) and S-SIRUS (blue dashed line) residuals, with associated bootstrap-based confidence regions (shadow area) for the three scenarios.}
      \centering
	   \begin{subfigure}{0.32\linewidth}
		\includegraphics[width=\linewidth]{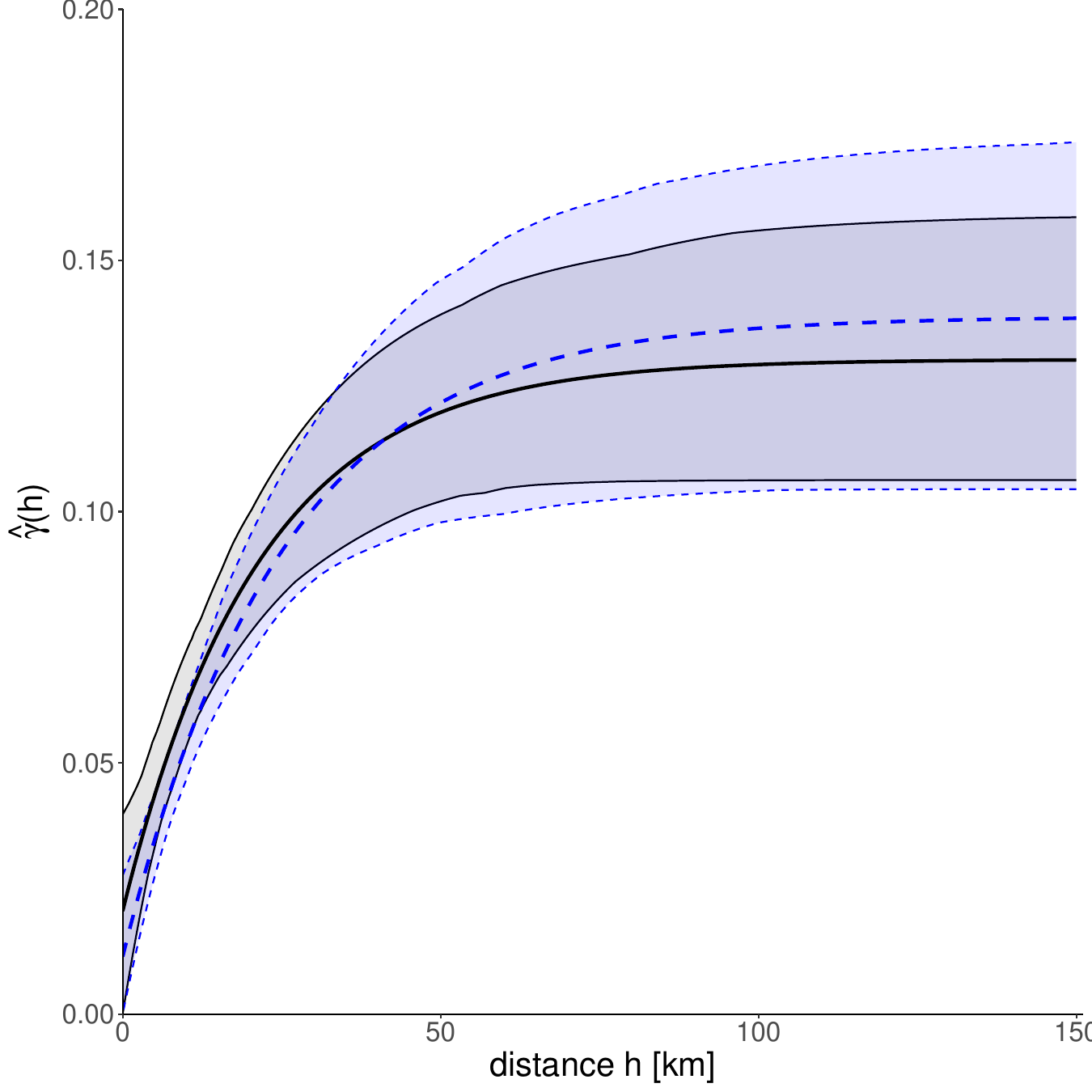}
		\caption{Scenario A}
	   \end{subfigure}
	   \begin{subfigure}{0.32\linewidth}
		\includegraphics[width=\linewidth]{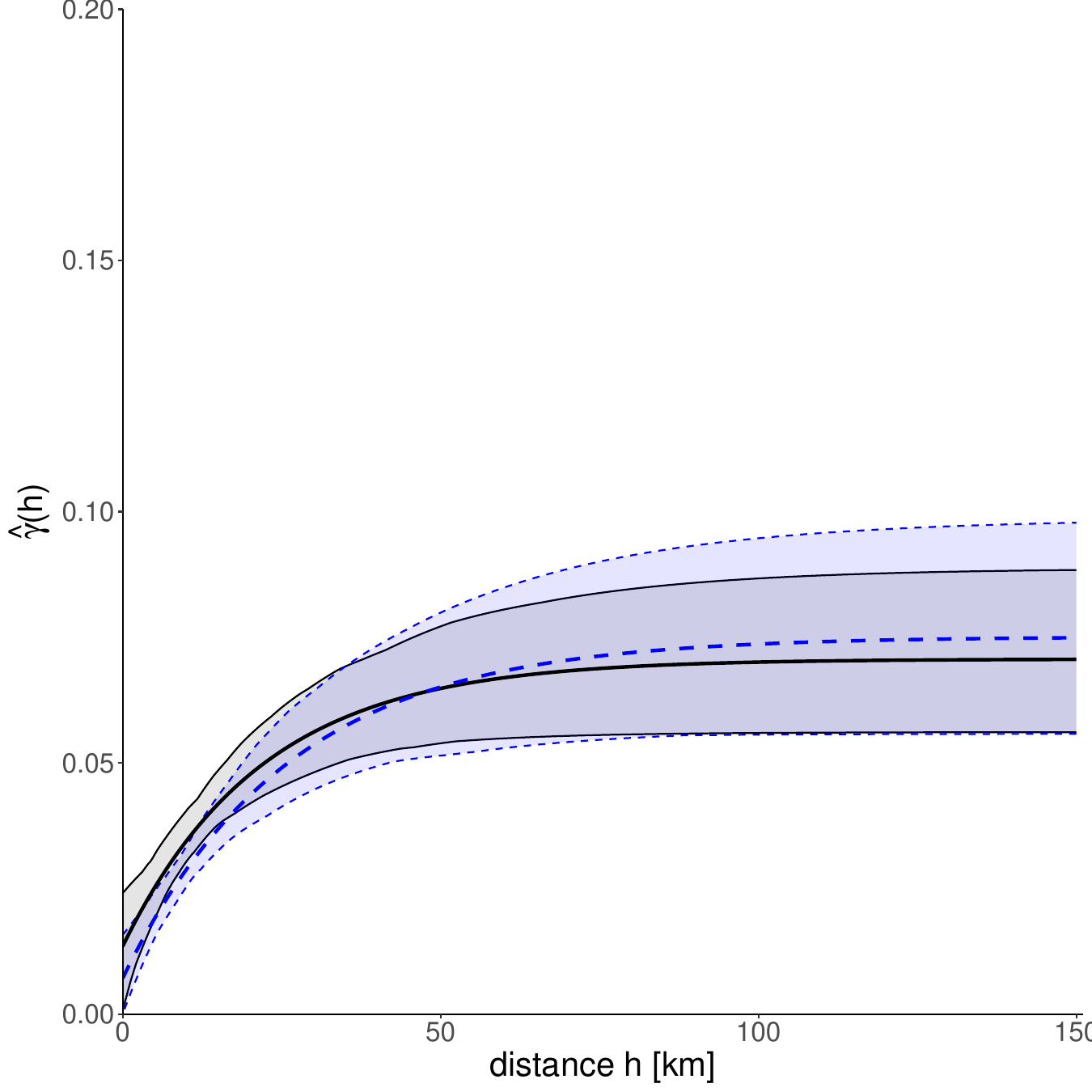}
		\caption{Scenario B}
	    \end{subfigure}
	     \begin{subfigure}{0.32\linewidth}
		 \includegraphics[width=\linewidth]{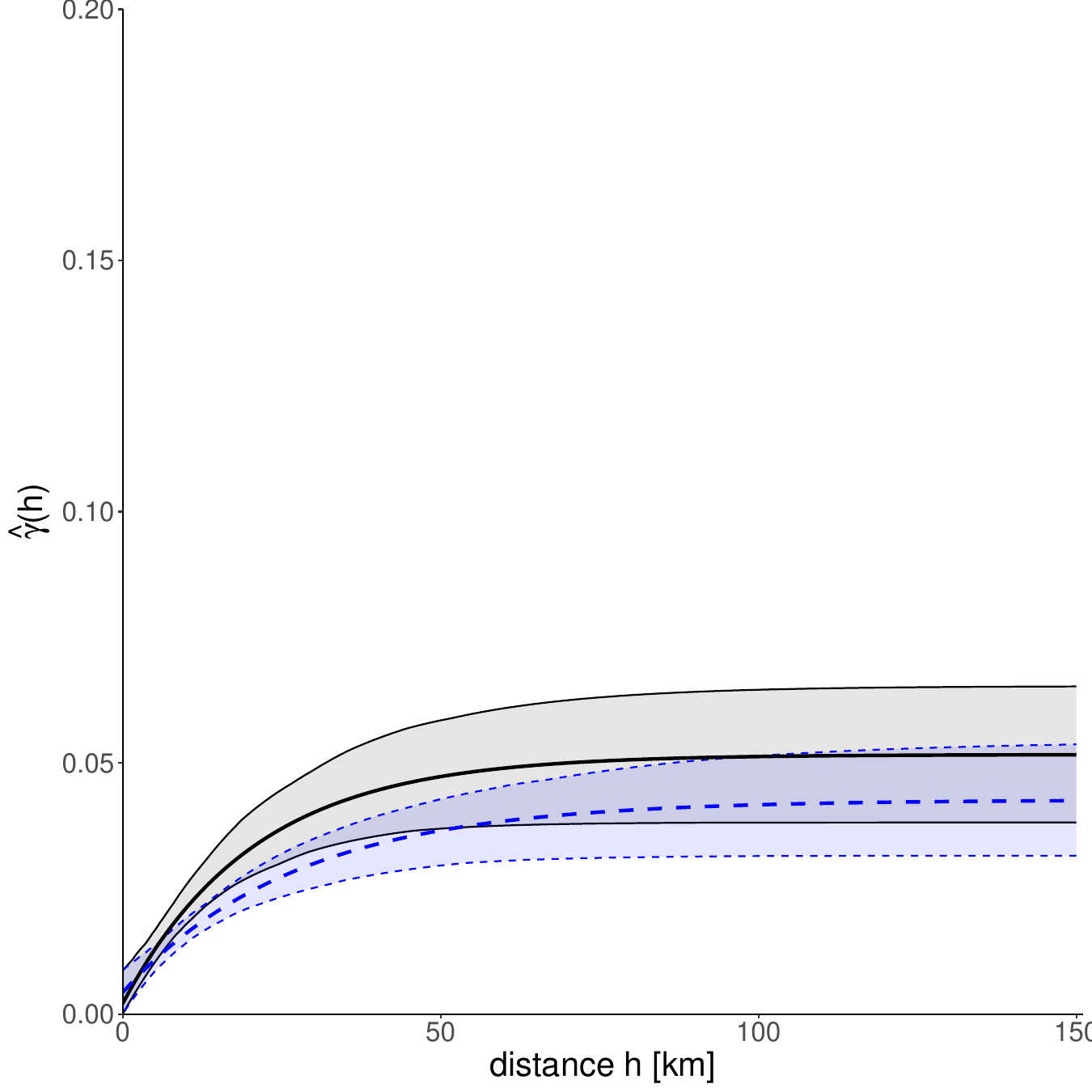}
		 \caption{Scenario C}
	      \end{subfigure}
	\label{fig:residualsvariograms}
\end{figure}

Predictions for the S-SIRUS and SIRUS large-scale components (according to Eq.~\eqref{Eq:mainmodel_prediction}) are obtained by recovering the corresponding rule value $\hat{g}_{n,\mathscr{P}}(\mathbf{x}(s_0))$ for each rule generated in the fitting phase.
As explained in Section~\ref{subsec:ssiruspredict}, to obtain response variable predictions for the new observations in $\mathscr{D}_{100}$ a post-processing strategy is employed, requiring the fit of an ordinary kriging model to the residuals.
To this goal, we consider the approach proposed by \cite{datta2016hierarchical} based on the Nearest Neighbour Gaussian Processes as detailed in \cite{saha2018brisc} and implemented in the \texttt{R} package \texttt{BRISC}.
Figure \ref{fig:residualsvariograms} illustrates the estimated exponential variograms $\hat \gamma(h)$ for SIRUS (black solid line) and S-SIRUS (blue dashed line) residuals, with corresponding confidence regions obtained by bootstrap, for the three scenarios.
Moving from the left to the right panel, we observe that the residual spatial structure becomes weaker in accordance with the data simulation assumptions about the SNR values. It is evident that a post-processing strategy is necessary, given the spatial correlation remaining in the residuals: an ordinary kriging model is fitted over the SIRUS and S-SIRUS residuals, i.e., residual kriging (RK) is considered. We refer to these two post-processing strategies as SIRUS-RK and S-SIRUS-RK.

Table~\ref{tab:prediction} provides the values of the test unexplained variance for SIRUS and S-SIRUS and their corresponding version with the ordinary kriging model fitted over the residuals obtained for Scenario A, B and C. We consider the test unexplained variance as a measure of the prediction error. For all four models, the best predictive performance is observed when the data exhibit less spatial correlation, i.e. in Scenario C. Furthermore, the results indicate that S-SIRUS models outperform the SIRUS ones in terms of predictive accuracy, suggesting that they are capable of effectively leveraging spatial correlation, even in the case when it is smaller.
\begin{table}[ht]
\caption{Unexplained variance computed using the test predictions for SIRUS, S-SIRUS, and their corresponding version with the ordinary kriging model fitted over the residuals (SIRUS-RK and S-SIRUS-RK).}
\begin{center}
\begin{tabular}[h]{lccc}
\toprule
 & \multicolumn{3}{c}{Test Unexplained Variance} \\
\cmidrule(lr{1em}){2-4}
\multirow{2}{*}{Model} & Scenario A & Scenario B & Scenario C \\
 & SNR = 0.5 & SNR = 1& SNR = 2\\
\midrule
SIRUS     & 0.61 & 0.53  & 0.44 \\        
S-SIRUS  & 0.56 & 0.47 & 0.32 \\
SIRUS-RK   & 0.33 & 0.31 & 0.24 \\
S-SIRUS-RK & 0.28   & 0.27 & 0.17 \\
\bottomrule
\label{tab:prediction}
\end{tabular}
\end{center}
\end{table}

%%%%%%%%%
\section{Conclusions}
\label{sec:conclusion}
The aim of our work is to propose an explainable RF algorithm which can be used when data are collected for spatial phenomena. In particular, we propose S-SIRUS, a spatially aware version of SIRUS, a rule extraction algorithm for regression problems originally proposed for independent data \citep{benard21regression}. By using S-SIRUS it is possible to obtain a stable and short list of rules from the RF-GLS algorithm, a regression RF algorithm specifically developed for spatially correlated data \citep{saha2023random}. However, we point out that S-SIRUS can estimate the large-scale component of the considered spatial model but is not directly able to predict and explain the response variable and then a post-processing strategy should be considered (e.g., S-SIRUS-RK).

The CV results from the simulation study show that S-SIRUS outperforms SIRUS in terms of simplicity and stability when spatial dependence among data exists while exhibiting a similar predictive accuracy. 
Instead, by using a test set to assess the predictive performance, we observe that S-SIRUS models outperform the SIRUS ones in terms of lower unexplained variance.

It is worth noting that, given the same number of trees, the computational cost for fitting S-SIRUS is always higher than that of SIRUS, given the higher computational load required to fit RF-GLS rather than standard RF. For example, in Scenario A, fitting $1000$ trees using the \texttt{sirus.fit} \texttt{R} function takes\footnote{The times were obtained using a computer equipped with an AMD Ryzen 5 3600 6-Core Processor 3.60 GHz (no core parallelisation) and 16 GB of RAM.}  $0.1455$ seconds, while using \texttt{s.sirus.fit} requires 
$1.6832$ minutes. Specifically, \texttt{sirus.fit} calls the \texttt{ranger} function from the homonym \texttt{R} package, taking 
$0.0168$ seconds to grow a forest with $1000$ trees, whereas \texttt{s.sirus.fit} calls \texttt{RFGLS$\_$estimate$\_$spatial} from the \texttt{RandomForestsGLS} package, which requires 
$43.7448$ seconds.  

To the best of our knowledge, S-SIRUS is the first rule algorithm for regression proposed in the literature to explain an RF for spatially correlated data. In this work, we have chosen RF-GLS as the most suitable black box model to explain. However, as future research, we plan to extend S-SIRUS to incorporate other techniques for adjusting regression RF algorithms for spatially correlated data, such as the method proposed by \cite{rabinowicz2022tree}. Another possible extension of S-SIRUS involves considering the bias correction CV version proposed by \cite{rabinowicz2022cross} to fine-tune  $p_0$ when a standard CV is unsuitable for spatially correlated data.

\clearpage

%%%%%%%%%
\appendix

\section{Output of the the \texttt{R} function \texttt{gam} from the \texttt{mgcv package} for the final GAM model \label{appendix:gam}}
\begin{verbatim}
# -------------------------------------------------------------------

Family: gaussian 
Link function: identity 

Formula:
log(AQ_pm10) ~ s(Altitude) + WE_temp_2m + WE_surface_pressure

Parametric coefficients:
                    Estimate Std. Error t value Pr(>|t|)    
(Intercept)          4.02051    0.01791 224.481  < 2e-16 ***
WE_temp_2m          -0.35528    0.03728  -9.531 1.46e-15 ***
WE_surface_pressure  0.59869    0.04755  12.590  < 2e-16 ***
---
Signif. codes:  0 ‘***’ 0.001 ‘**’ 0.01 ‘*’ 0.05 ‘.’ 0.1 ‘ ’ 1

Approximate significance of smooth terms:
              edf Ref.df     F p-value    
s(Altitude) 6.651  7.658 28.09  <2e-16 ***
---
Signif. codes:  0 ‘***’ 0.001 ‘**’ 0.01 ‘*’ 0.05 ‘.’ 0.1 ‘ ’ 1

R-sq.(adj) =  0.879   Deviance explained = 88.9%
GCV = 0.037408  Scale est. = 0.034002  n = 106

# -------------------------------------------------------------------
\end{verbatim}

\newpage

\section{SIRUS and S-SIRUS rules for scenario B and C}
\label{appendix:furtherresults}

\begin{figure}[!ht]
\caption{Unexplained variance (top panel) and stability (bottom panel) versus the number of rules for SIRUS (left panels) and S-SIRUS (right panels) in Scenario B (SNR = 1) for a fine grid of $p_0$, assessed via standard CV with $K=10$ folds. The results are averaged and bars show the variability of the metrics across 10 repetitions. 
The optimal $p_0$ value is the median $p_0$ value across the 10 CV repetitions (0.0257 for SIRUS and 0.0232 for S-SIRUS).
}
      \centering
	   \begin{subfigure}{0.45\linewidth}
		\includegraphics[width=\linewidth]{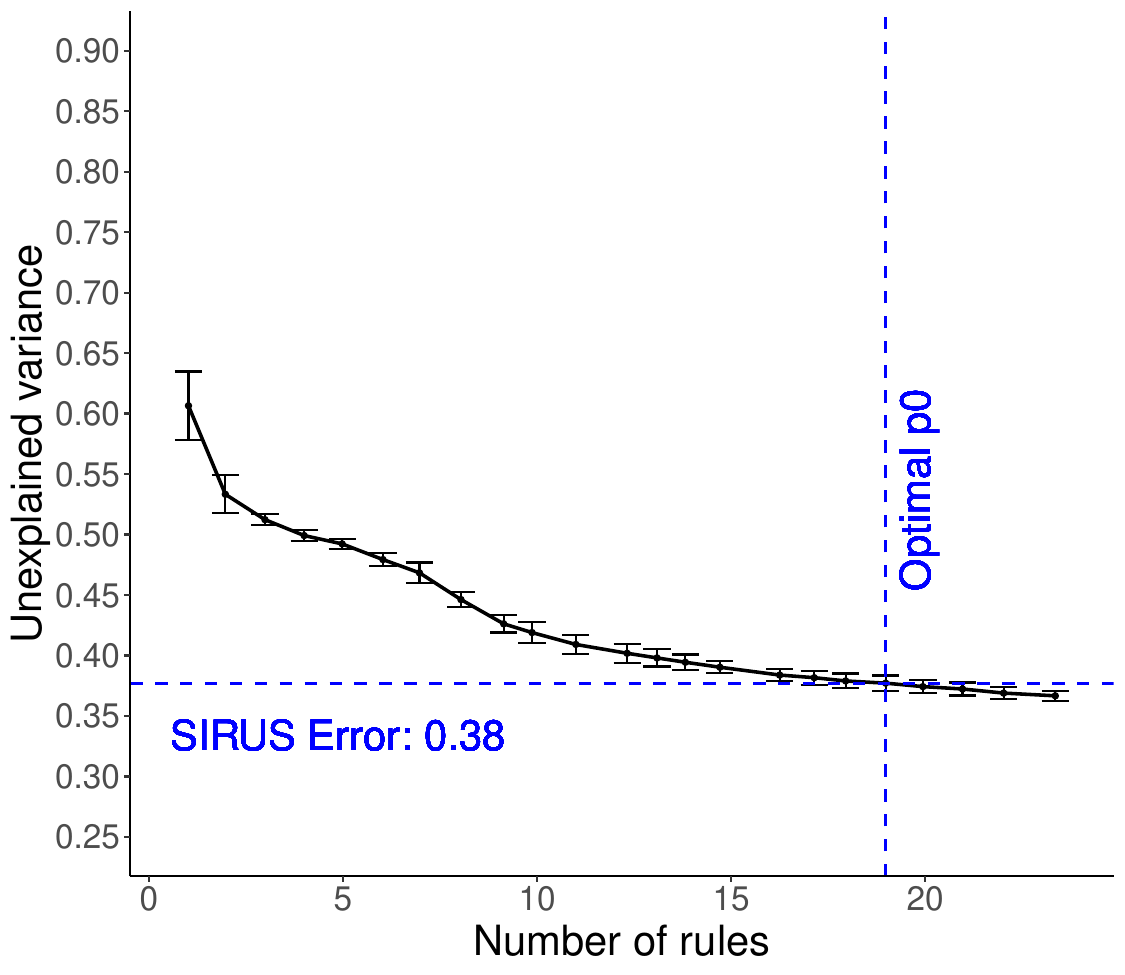}
		\caption{Unexplained variance SIRUS}
	   \end{subfigure}
	   \begin{subfigure}{0.45\linewidth}
		\includegraphics[width=\linewidth]{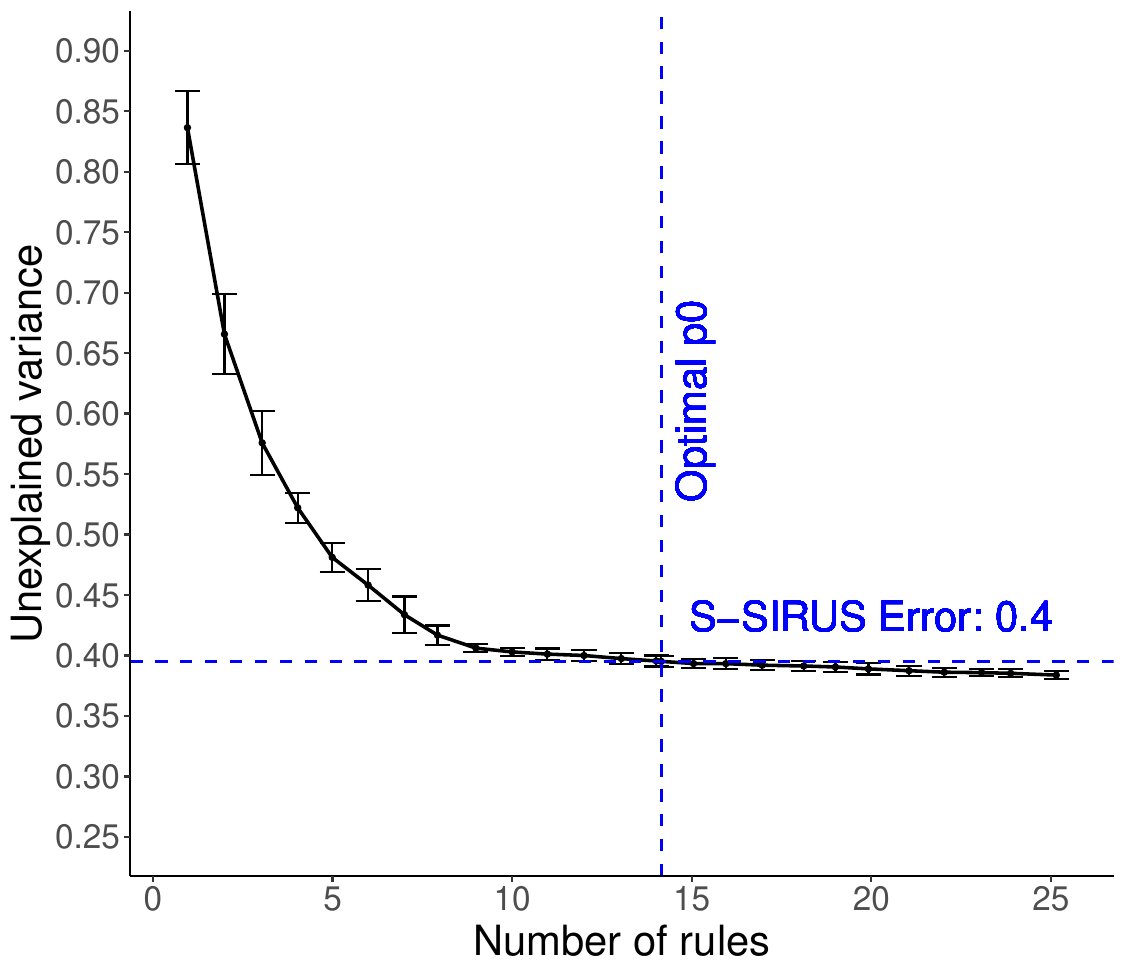}
		\caption{Unexplained variance S-SIRUS}
	    \end{subfigure}
	\vfill
	     \begin{subfigure}{0.45\linewidth}
		 \includegraphics[width=\linewidth]{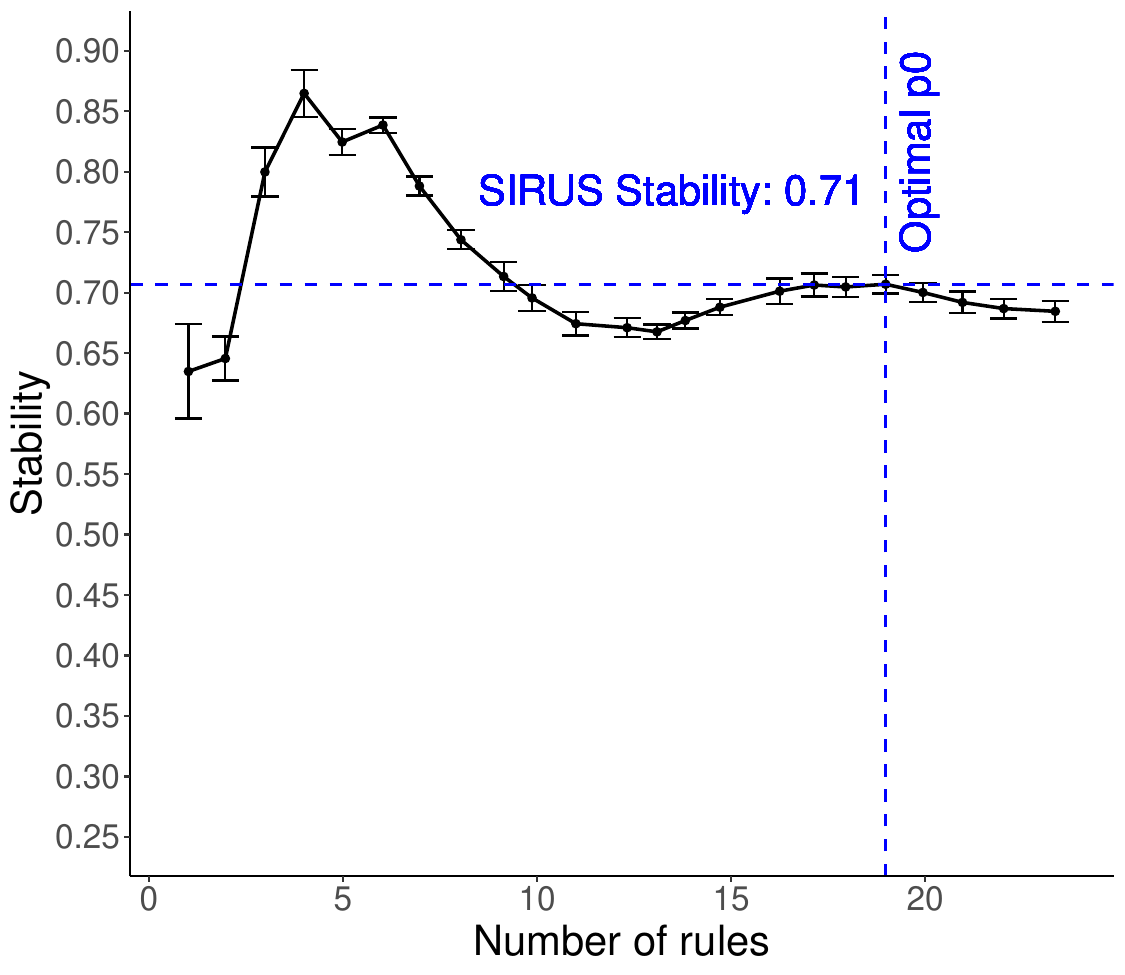}
		 \caption{Stability SIRUS}
	      \end{subfigure}
	       \begin{subfigure}{0.45\linewidth}
		  \includegraphics[width=\linewidth]{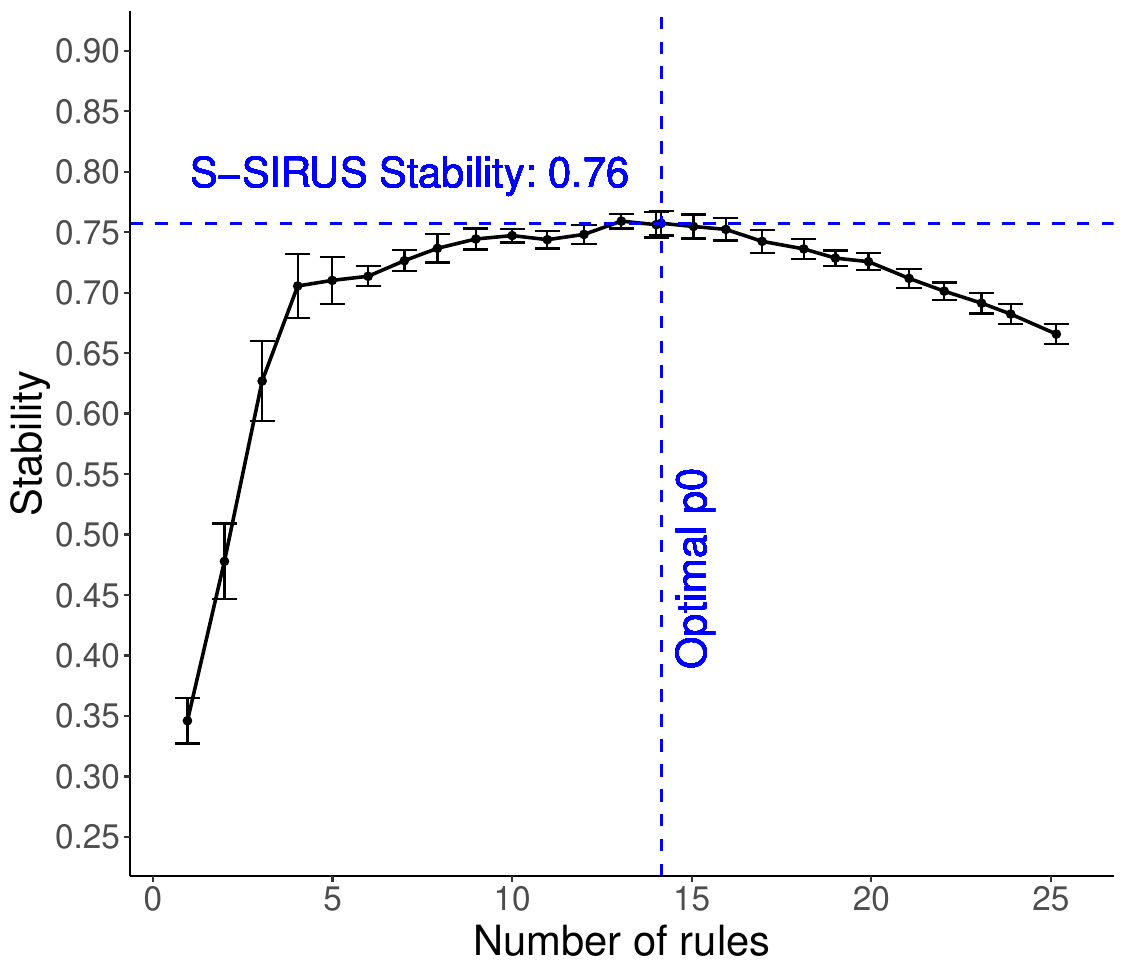}
		  \caption{Stability S-SIRUS}
	       \end{subfigure}
	\label{fig:cvfunctionoutB}
\end{figure}

\begin{landscape}
\begin{table}
    \centering
    \caption{SIRUS (top) and S-SIRUS (bottom) list of rules for Scenario B (SNR = 0.5). To reach convergence, $11000$ trees are grown for SIRUS and 14000 for S-SIRUS.
    }
    \scriptsize{
    \begin{tabular}[t]{llc}
    \toprule
\multicolumn{1}{c}{\textbf{Average}  $\log$(PM$_{10}$) =  4.04 } & \multicolumn{1}{c}{\textbf{Intercept} = -4.46} & \textbf{n} = 400\\
\\
\multicolumn{1}{l}{\textbf{Weight}} & \multicolumn{1}{c}{\textbf{SIRUS Rule}} & \multicolumn{1}{c}{\textbf{Frequency}} \\
\midrule
\rowcolor{gray!20}
0.18 & if solar\_radiation $<$ 0.66 then 3.73 (n=200) else 4.36 (n=200) & 0.235\\
0.0999 & if surface\_pressure $<$ 0.286 then 3.74 (n=200) else 4.34 (n=200) & 0.155  \\
\rowcolor{gray!20}
0.0953 & if rh\_mean $<$ -0.275 then 3.69 (n=160) else 4.28 (n=240) & 0.152 \\
\rowcolor{gray!20}
0.0416 & if blh\_layer\_max $<$ -0.173 then 3.63 (n=120) else 4.22 (n=280) & 0.075  \\
0.134 & if solar\_radiation $<$ 0.66 \& surface\_pressure $<$ -1.46 then 3.34 (n=40) else 4.12 (n=360) & 0.060 \\
0.0607 & if Altitude $<$ -0.332 then 4.33 (n=200) else 3.76 (n=200) & 0.059 \\
\rowcolor{gray!20}
0.033 & if surface\_pressure $<$ -0.122 then 3.68 (n=160) else 4.29 (n=240) & 0.055 \\
0.143 & if solar\_radiation $\ge$ 0.66 \& surface\_pressure $\ge$ 0.286 then 4.38 (n=181) else 3.77 (n=219) & 0.041 \\
0.0408 & if rh\_mean $\ge$ -0.275 \& wind\_speed\_100m\_mean $<$ -0.306 then 3.97 (n=54) else 4.06 (n=346) & 0.040 \\
0.144 & if temp\_2m $\ge$ 0.329 \& wind\_speed\_100m\_mean $<$ -0.306 then 3.9 (n=51) else 4.07 (n=349) & 0.039 \\
\rowcolor{gray!20}
0.19 & if Altitude $<$ 1.17 then 4.18 (n=320) else 3.49 (n=80) & 0.038 \\
0.00295 & if surface\_pressure $\ge$ 0.286 \& wind\_speed\_100m\_mean $<$ -0.306 then 4.04 (n=27) else 4.04 (n=373) & 0.033 \\
0.173 & if surface\_pressure $\ge$ 0.286 \& nox\_sum $<$ 0.0438 then 4.42 (n=131) else 3.86 (n=269) & 0.033 \\
0.107 & if temp\_2m $<$ -1.56 \& rh\_mean $<$ -0.275 then 3.35 (n=37) else 4.11 (n=363) & 0.032 \\
0.00147 & if rh\_mean $\ge$ -0.275 \& solar\_radiation $<$ 0.66 then 3.97 (n=44) else 4.05 (n=356) &0.032  \\
0.134 & if temp\_2m $<$ 0.329 \& surface\_pressure $<$ -1.46 then 3.34 (n=40) else 4.12 (n=360) & 0.029 \\
0.0278 & if temp\_2m $\ge$ 0.329 \& surface\_pressure $<$ 0.286 then 3.95 (n=40) else 4.05 (n=360) & 0.029 \\
0.147 & if solar\_radiation $\ge$ 0.66 \& nox\_sum $<$ 0.0438 then 4.4 (n=140) else 3.85 (n=260)
 & 0.029 \\
0.125 & if blh\_layer\_max $<$ 0.18 \& solar\_radiation $\ge$ 0.66 then 4.56 (n=18) else 4.02 (n=382) & 0.028\\
0.222 & if solar\_radiation $<$ 0.66 \& nox\_sum $<$ -0.444 then 3.38 (n=38) else 4.11 (n=362)
 & 0.028\\
\bottomrule
 & & \\
\end{tabular}
\begin{tabular}[t]{llc}
\toprule
\multicolumn{1}{c}{\textbf{Average}  $\log$(PM$_{10}$) =  4.04}  & \multicolumn{1}{c}{\textbf{Intercept} = -3.04} & \textbf{n} = 400\\
\\
\multicolumn{1}{l}{\textbf{Weight}} & \multicolumn{1}{c}{\textbf{S-SIRUS Rule}} & \multicolumn{1}{c}{\textbf{Frequency}} \\
\midrule
0.151 & if nox\_sum $<$ -0.444 then 3.42 (n=40) else 4.11 (n=360) & 0.122 \\
0.0511 & if surface\_pressure $<$ -1.11 then 3.52 (n=80) else 4.18 (n=320)& 0.102  \\
0.253 & if Altitude $<$ 1.53 then 4.13 (n=360) else 3.28 (n=40) & 0.091 \\
\rowcolor{gray!20}
0.00603 & if blh\_layer\_max $<$ -0.173 then 3.63 (n=120) else 4.22 (n=280) &0.087  \\
0.133 & if rh\_mean $<$ -1.04 then 3.52 (n=80) else 4.18 (n=320)& 0.069 \\
\rowcolor{gray!20}
0.35 & if solar\_radiation $<$ 0.66 then 3.73 (n=200) else 4.36 (n=200)& 0.058 \\
0.0943 & if solar\_radiation $<$ -1.52 then 3.39 (n=40) else 4.12 (n=360) & 0.045\\
\rowcolor{gray!20}
0.149 & if rh\_mean $<$ -0.275 then 3.69 (n=160) else 4.28 (n=240)& 0.044 \\
\rowcolor{gray!20}
0.138 & if Altitude $<$ 1.17 then 4.18 (n=320) else 3.49 (n=80)& 0.040 \\
0.192 & if wind\_speed\_100m\_mean $<$ -0.306 then 3.8 (n=160) else 4.21 (n=240)&0.036 \\
0.0246 & if nox\_sum $<$ -0.406 then 3.62 (n=80) else 4.15 (n=320) &0.033 \\
\rowcolor{gray!20}
0.0942 & if surface\_pressure $<$ -0.122 then 3.68 (n=160) else 4.29 (n=240) &0.027 \\
0.0394 & if Altitude $<$ 0.00971 then 4.27 (n=240) else 3.7 (n=160) &0.024 \\
0.0764 & if temp\_2m $<$ -1.56 then 3.4 (n=40) else 4.12 (n=360) & 0.024\\
\bottomrule
\end{tabular}
}
\label{tab:rulesscenarioB}
\end{table}
\end{landscape}

\begin{figure}
\caption{Unexplained variance (top panel) and stability (bottom panel) versus the number of rules for SIRUS (left panels) and S-SIRUS (right panels) in Scenario C (SNR = 2) for a fine grid of $p_0$, assessed via standard CV with $K=10$ folds. The results are averaged and bars show the variability of the metrics across 10 repetitions.
The optimal $p_0$ value is the median $p_0$ value across the 10 CV repetitions (0.0616 for SIRUS and 0.0196 for S-SIRUS).
}

      \centering
	   \begin{subfigure}{0.495\linewidth}
		\includegraphics[width=\linewidth]{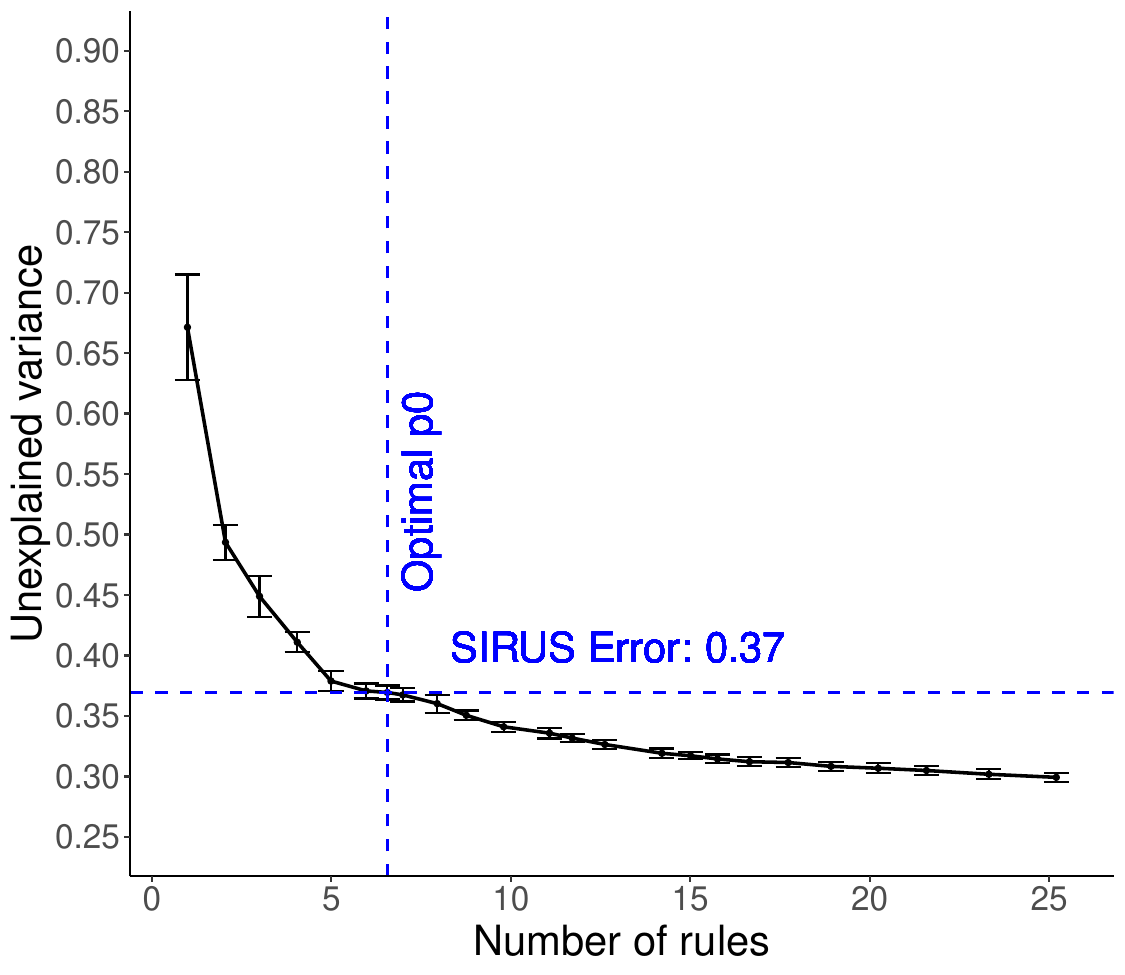}
		\caption{Unexplained variance SIRUS}
	   \end{subfigure}
	   \begin{subfigure}{0.495\linewidth}
		\includegraphics[width=\linewidth]{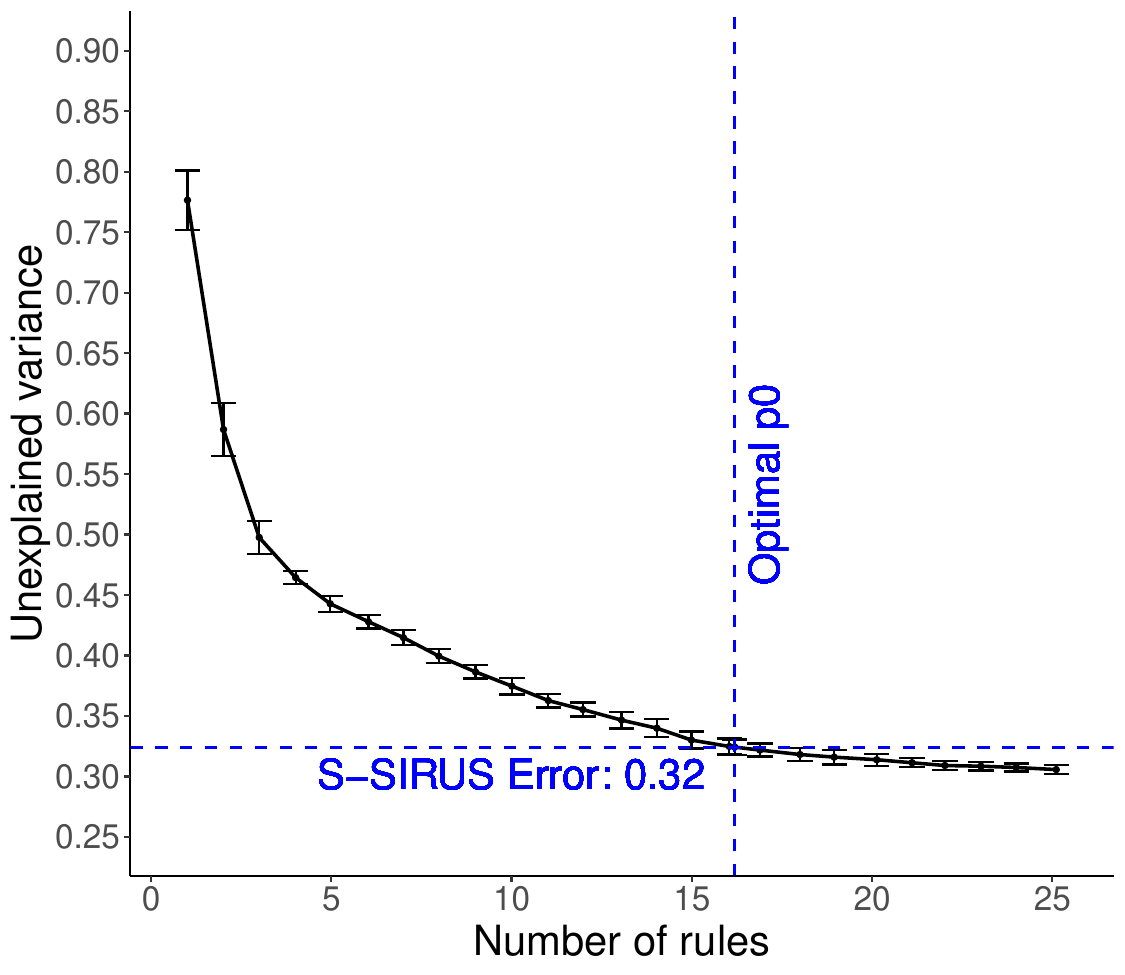}
		\caption{Unexplained variance S-SIRUS}
	    \end{subfigure}
	\vfill
	     \begin{subfigure}{0.495\linewidth}
		 \includegraphics[width=\linewidth]{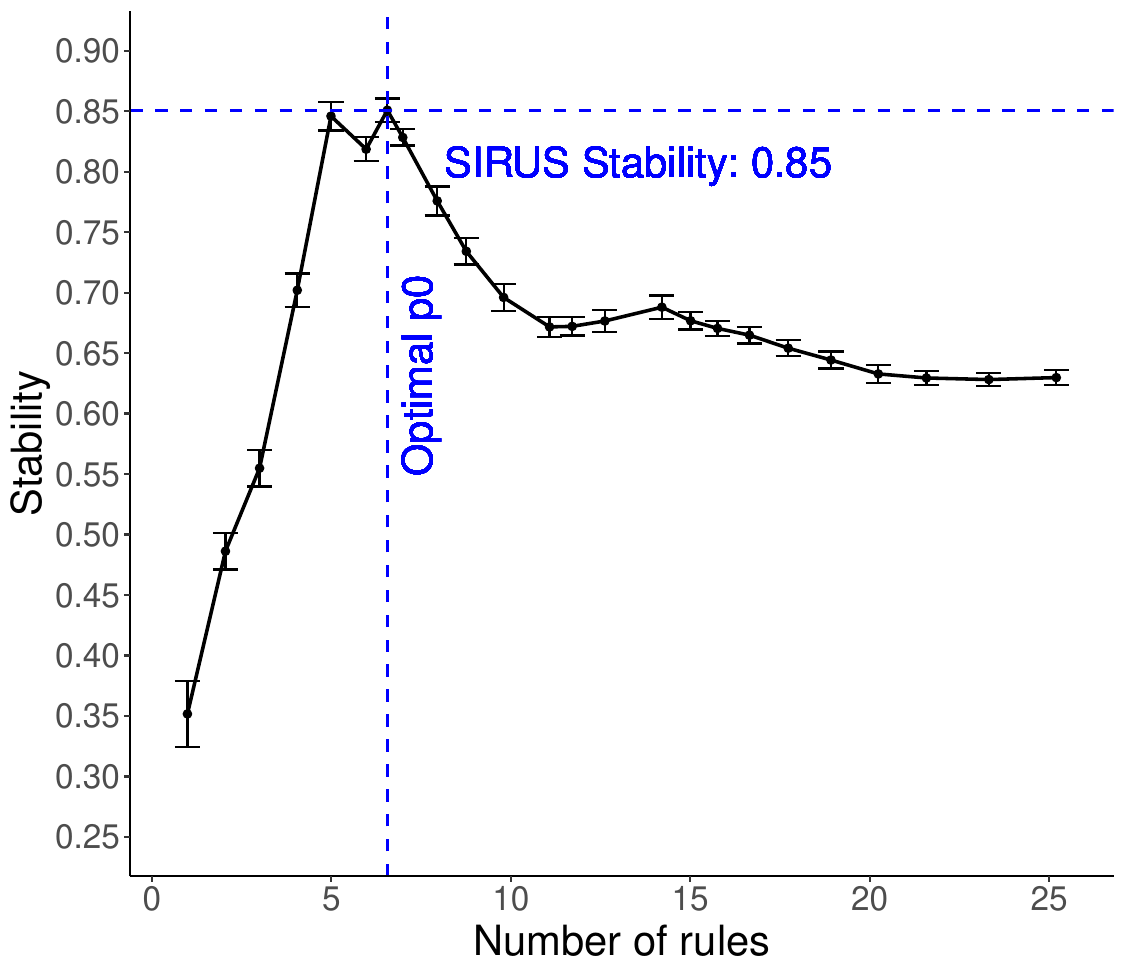}
		 \caption{Stability SIRUS}
	      \end{subfigure}
	       \begin{subfigure}{0.495\linewidth}
		  \includegraphics[width=\linewidth]{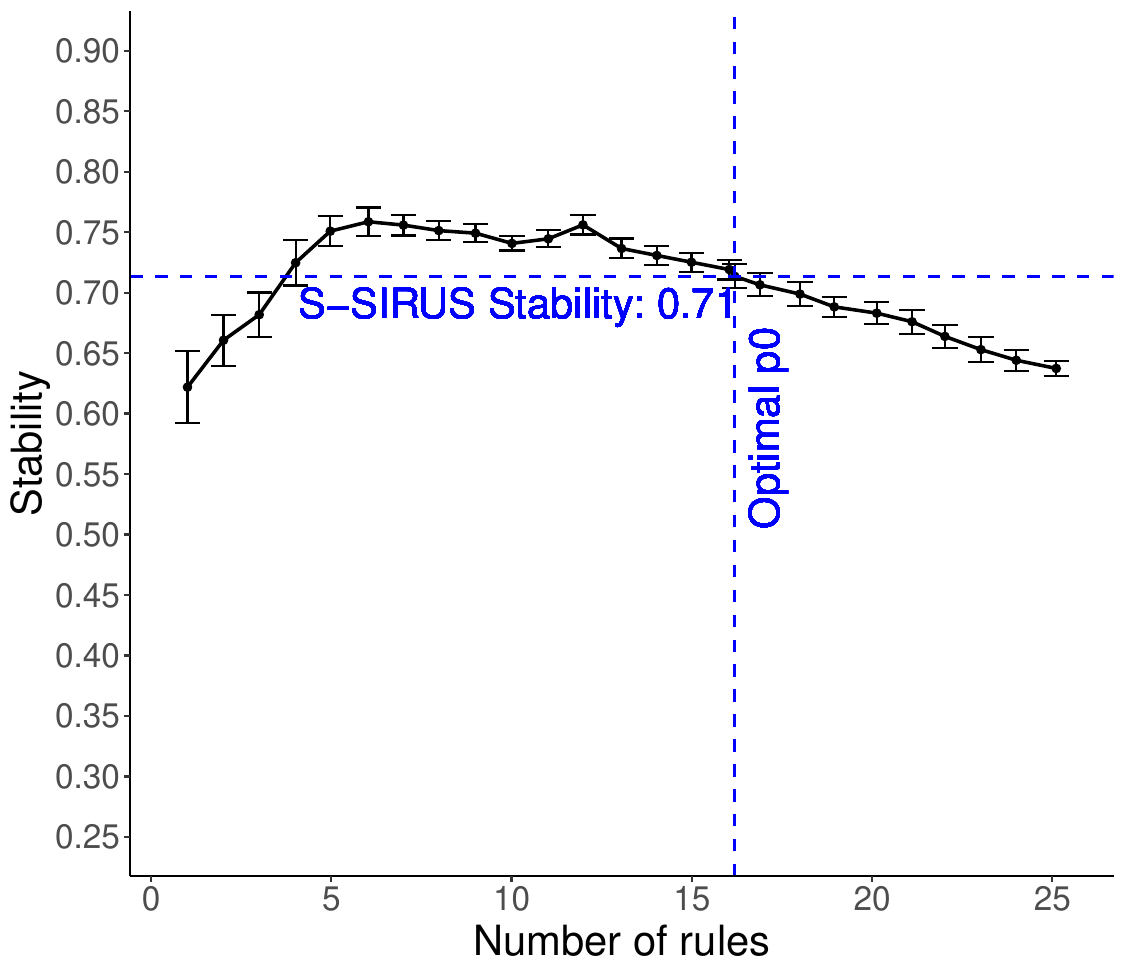}
		  \caption{Stability S-SIRUS}
	       \end{subfigure}
	\label{fig:cvfunctionoutC}
\end{figure}
\begin{landscape}
\begin{table}
    \centering
    \caption{SIRUS (top) and S-SIRUS (bottom) list of rules for Scenario C (SNR = 2). To reach convergence, $12000$ trees are grown for SIRUS and 17000 for S-SIRUS.
}
    \scriptsize{
    \begin{tabular}[t]{llc}
\toprule
\multicolumn{1}{c}{\textbf{Average}  $\log$(PM$_{10}$)} =  4.02  & \multicolumn{1}{c}{\textbf{Intercept} = -1.16} & \textbf{n} = 400\\
\\
\multicolumn{1}{l}{\textbf{Weight}} & \multicolumn{1}{c}{\textbf{SIRUS Rule}} & \multicolumn{1}{c}{\textbf{Frequency}} \\
\midrule
\rowcolor{gray!20}
 0.277 & if solar\_radiation $<$ 0.66 then 3.75 (n=200) else 4.29 (n=200) & 0.158 \\
 0.283 & if surface\_pressure $<$ 0.286 then 3.75 (n=200) else 4.29 (n=200) & 0.152  \\
 \rowcolor{gray!20}
 0.174 & if rh\_mean $<$ -0.275 then 3.7 (n=160) else 4.24 (n=240)& 0.144 \\
 \rowcolor{gray!20}
 0.418 & if Altitude $<$ 1.17 then 4.15 (n=320) else 3.49 (n=80) & 0.129 \\
 0.0155 & if temp\_2m $<$ 0.329 then 3.7 (n=160) else 4.23 (n=240) & 0.124 \\
 \rowcolor{gray!20}
 0.122 & if blh\_layer\_max $<$ -0.173 then 3.64 (n=120) else 4.18 (n=280)& 0.071\\
\bottomrule
 & & \\
\end{tabular}

\begin{tabular}[t]{llc}
\toprule
\multicolumn{1}{c}{\textbf{Average}  $\log$(PM$_{10}$)} =  4.02 & \multicolumn{1}{c}{\textbf{Intercept} = -3.4} & \textbf{n} = 400\\
\\
\multicolumn{1}{l}{\textbf{Weight}} & \multicolumn{1}{c}{\textbf{S-SIRUS Rule}} & \multicolumn{1}{c}{\textbf{Frequency}} \\
\midrule
0.113 & if nox\_sum $<$ -0.444 then 3.43 (n=40) else 4.09 (n=360) & 0.152 \\
0.0774 & if surface\_pressure $<$ -1.11 then 3.51 (n=80) else 4.15 (n=320) &  0.123 \\
\rowcolor{gray!20}
0.031 & if blh\_layer\_max $<$ -0.173 then 3.64 (n=120) else 4.18 (n=280) & 0.098 \\
0.207 & if Altitude $<$ 1.53 then 4.1 (n=360) else 3.29 (n=40) & 0.095 \\
0.128 & if rh\_mean $<$ -1.04 then 3.53 (n=80) else 4.14 (n=320) & 0.088 \\
0.0472 & if solar\_radiation $<$ -1.52 then 3.39 (n=40) else 4.09 (n=360) & 0.075 \\
\rowcolor{gray!20}
0.161 & if Altitude $<$ 1.17 then 4.15 (n=320) else 3.49 (n=80) & 0.043 \\
\rowcolor{gray!20}
0.193 & if rh\_mean $<$ -0.275 then 3.7 (n=160) else 4.24 (n=240) & 0.032 \\
0.135 & if surface\_pressure $<$ -1.46 then 3.33 (n=40) else 4.1 (n=360) & 0.030 \\
0.0116 & if temp\_2m $<$ -1.56 then 3.41 (n=40) else 4.09 (n=360) & 0.027 \\
\rowcolor{gray!20}
0.335 & if solar\_radiation $<$ 0.66 then 3.75 (n=200) else 4.29 (n=200) & 0.023 \\
0.0118 & if blh\_layer\_max $<$ -0.173 \& temp\_2m $<$ -1.56 then 3.41 (n=40) else 4.09 (n=360)&0.023\\
0.177 & if wind\_speed\_100m\_mean $<$ -0.306 then 3.81 (n=160) else 4.16 (n=240) &0.021 \\
0.0757 & if nox\_sum $<$ -0.444 \& rh\_mean $<$ -1.37 then 3.21 (n=15) else 4.05 (n=385) & 0.020  \\
0.142 & if nox\_sum $<$ -0.444 \& solar\_radiation $<$ -1.52 then 3.13 (n=17) else 4.06 (n=383)&0.020\\
\bottomrule
\end{tabular}
}
\label{tab:rulesscenarioC}
\end{table}
\end{landscape}

%%%%%%%%%
\bibliographystyle{elsarticle-harv} 
\bibliography{ssiruspaper.bib}

\end{document}